\documentclass[final]{clv2025}

\jvol{}
\jnum{}
\jyear{}

\dochead{}

\pageonefooter{}

\usepackage{amsmath}
\usepackage{booktabs}

\usepackage{pifont}  
\newcommand{\cmark}{\textcolor{green!60!black}{\ding{51}}} 
\newcommand{\xmark}{\textcolor{red}{\ding{55}}}     

\usepackage{graphicx}
\usepackage{multirow}
\usepackage{times}
\usepackage{latexsym}
\usepackage{multirow}
\usepackage{amsmath}
\usepackage{amssymb}
\usepackage{amsthm}
\usepackage{amsfonts}
\usepackage{graphicx}
\usepackage{booktabs}
\usepackage{pdfpages}
\usepackage[table]{xcolor}
\usepackage[normalem]{ulem}
\useunder{\uline}{\ul}{}

\usepackage{subcaption}

\runningtitle{CACARA}
\runningauthor{Moreira et al.}

\title{CACARA: Cross-Modal Alignment Leveraging a Text-Centric Approach for Cost-Effective Multimodal and Multilingual Learning}

\author{Diego A.~B.~Moreira\thanks{Corresponding authors: \email{d230640@dac.unicamp.br}, \email{alef\_iury\_c.c@discente.ufg.br}}$^{,1}$, Alef I.~Ferreira$^{3}$, Jhessica Silva$^{1}$, Gabriel O.~dos Santos$^{1}$, Gustavo Bonil$^{2}$, João Gondim$^{1}$, Marina dos Santos$^{2}$, Helena Maia$^{1}$, Simone Hashiguti$^{2}$, Nádia da Silva$^{3}$, Carolina Scarton$^{4}$, Helio Pedrini$^{1}$, Sandra Avila$^{1}$}

\affilblock{
    \affil{Instituto de Computação, Universidade Estadual de Campinas (UNICAMP), Brasil}
    \affil{Instituto de Estudos da Linguagem, Universidade Estadual de Campinas (UNICAMP), Brasil}
    \affil{Instituto de Informática, Universidade Federal de Goiás (UFG), Goiás, Brasil}
    \affil{Department of Computer Science, University of Sheffield, Sheffield, United Kingdom}
}

\begin{document}

\maketitle

\begin{abstract}

\noindent As deep learning models evolve, new applications and challenges are rapidly emerging. Tasks that once relied on a single modality, such as text, images, or audio, are now enriched by seamless interactions between multimodal data. These connections bridge information gaps: an image can visually materialize a text, while audio can add context to an image. Researchers have developed numerous multimodal models, but most rely on resource-intensive training across multiple modalities. Similarly, extending these models to new languages often follows the same resource-heavy training strategy. In this work, we propose a multimodal and multilingual architecture, CACARA, trained through emergent alignment learning, enabling the seamless integration of new modalities into an existing bimodal/multimodal model without requiring full retraining. This work breaks new ground by demonstrating that this emergent alignment paradigm can unlock multilingual capabilities from monolingual training. By fine-tuning the newly incorporated modality only on data aligned with the English language, our model develops support for over 100 languages without explicit multilingual pretraining or tuning of the text encoder. Such emergent multimodal and multilingual properties are gained efficiently, preserving previously learned knowledge at a training cost comparable to that of a monolingual model. Our strategy achieves up to a 14.24 percentage points improvement in R@1 audio-to-text retrieval, outperforming state-of-the-art multimodal models—all without the heavy computational cost of retraining across every modality and language.
\end{abstract}
\section{Introduction}
\label{sec:intro}

Deep learning has revolutionized multiple domains by enabling models to learn complex representations across diverse data types. Early breakthroughs in computer vision, driven by convolutional neural networks~\cite{krizhevsky2012imagenet}, were followed by advances in natural language processing, culminating in Transformer networks~\cite{vaswani2017attention}. Beyond images and text, deep learning has achieved state-of-the-art performance in audio~\cite{vandenoord16_ssw}, sensor data~\cite{wang2019deep}, and tabular data~\cite{arik2021tabnet}, excelling in classification, retrieval, and generation tasks. 

Real-world applications, however, often involve complex interactions between multiple data types. For instance, video understanding encompasses the joint processing of visual and auditory information~\cite{goecke2005current}. These models integrate complementary information from different modalities, as exemplified by CLIP (Contrastive Language-Image Pretraining)~\cite{DBLP:journals/corr/abs-2103-00020}, which learns a joint representation space for images and text, enabling cross-modal retrieval and zero-shot classification. The benefits of multimodal learning extend beyond simple fusion, uncovering latent relationships and contextual cues that are not apparent in individual modalities~\cite{baltruvsaitis2018multimodal}. However, training such models is challenging due to the need for synchronized data and the high cost of annotated datasets. One promising approach is implicit learning, where the model implicitly learns cross-modal relationships, even without strict temporal alignment, by leveraging the inherent correlations and statistical dependencies between modalities~\cite{alayrac2020self}.

Multilingualism adds another layer of complexity. Supporting multiple languages not only expands accessibility but also enriches models with diverse linguistic structures~\cite{conneau-etal-2020-unsupervised}. However, medium/low-resource languages remain underrepresented due to data scarcity and limited computational resources~\cite{joshi-etal-2020-state}. Current research disproportionately favors high-resource languages such as English, neglecting the needs of under-represented linguistic communities. 

The intersection of multimodality and multilingualism presents both opportunities and challenges. A key concern is the computational cost of training and deploying large-scale models, restricting access to well-resourced institutions~\cite{strubell-etal-2019-energy, strubell2020energy}. Therefore, there is a pressing need for innovative training methodologies and model architectures that can effectively leverage multimodal and multilingual data while minimizing computational~overhead. 

In this work, we introduce a multimodal and multilingual model that addresses these challenges through two key strategies: emergent alignment learning and a modified Locked-image Text Tuning (LiT)~\cite{zhai2022lit} protocol. These strategies reduce training costs while preserving high performance. 
We develop a new modality integration approach that eliminates the need to retrain all encoders. By optimizing this alignment with English-only data, we demonstrate that emergent alignment also benefits other languages. This aspect is not addressed in prior work such as~\citet{girdhar2023imagebind} and~\citet{zhulanguagebind}. We fine-tune the audio encoder for English synchronization only to enable multilingual capabilities without incurring the high costs of multilingual audio pretraining. The text encoder, meanwhile, remains frozen throughout this process, capitalizing on its inherent cross-lingual capabilities. This enables multilingual audio-text alignment in languages beyond English with training costs comparable to a monolingual model, unlike the strategy adopted in works such as~\citet{chen2023vast} and~\citet{zhang2023multimodal}, which incur high training costs due to the volume of data required for their multimodal or multilingual components, respectively.

Our findings show that multimodal models can learn language-agnostic concepts, improving R@1 text retrieval with audio by up to $14.24$ percentage points (pp) and audio-to-text retrieval by $2.58$ pp over existing multimodal approaches. Additionally, our method shows how to extend bilingual or multimodal models into a multilingual framework with minimal computational overhead while maintaining performance across modalities. It achieves an average classification accuracy of up to $66.5\%$ across multiple languages without requiring retraining or explicit alignment. These results demonstrate a scalable approach for efficiently integrating multiple modalities and languages.


The key contributions of this work are: (1) we propose a multimodal learning strategy based on emergent alignment and a modified Locked-image Tuning (LiT) protocol to seamlessly incorporate a new modality into an existing multimodal model; (2) we show that training only the newly added encoder against the text encoder is sufficient to implicitly align it with the shared multimodal space of all modalities; (3) we demonstrate that this emergent alignment enables zero-shot handling of previously untrained modality-modality feature pairs without any additional cross-modal training; (4) we empirically show that these strategies substantially reduce training costs while preserving high performance; (5) we extend emergent alignment to multilingual learning by optimizing audio-text alignment using English-only data, which, to the best of our knowledge, is the first such demonstration in the literature; and (6) we enable the audio component to support over 100 languages without explicit multilingual audio pretraining or retraining the text encoder, keeping the overall training cost comparable to that of a monolingual~model.

\section{Related Work}
\label{sec:related_work}

Multimodal learning has expanded machine learning’s scope, enabling models to process diverse data types. Foundational works like CLIP aligned images and text, inspiring extensions to other modalities such as audio, depth, and multilingual applications. \textbf{CAPIVARA}~\cite{santos2023capivara}, a CLIP-based model, incorporates Portuguese in contrastive training to optimize performance in low-resource languages. Despite progress, challenges remain in efficient training and generalization, particularly in low-resource settings. This section reviews advances in multimodal and multilingual~models.

\textbf{ImageBind}~\cite{girdhar2023imagebind} extends CLIP’s paradigm by introducing a unified embedding space for six modalities: images, text, audio, depth, thermal, and Inertial Measurement Unit (IMU) data. By leveraging contrastive learning and using images as an anchor modality, ImageBind showed that modalities can be effectively aligned through their natural pairing with images, eliminating the need for exhaustive paired data between all modality combinations. This approach achieves emergent cross-modal alignment without explicit supervision, demonstrating strong zero-shot transfer, enabling cross-modal retrieval and multimodal embedding arithmetic.

In addition to the natural alignment strategy proposed by ImageBind, the decision to use images as the anchor modality introduces certain limitations. As noted by~\citet{zhulanguagebind} as the backbone architecture for all modalities (except text), rather than utilizing modality-specific pretrained encoders. This homogeneous architectural choice may underutilize the unique representational strengths of different data types, potentially limiting performance when integrating new modalities.

\textbf{LanguageBind}~\cite{zhulanguagebind} replaces images with language as the central modality for aligning different data types. Leveraging language’s rich semantic structure, it aligns modalities within a shared embedding space using a frozen language encoder pretrained on video-language data and contrastive learning for other modalities. Efficient training is achieved through Low-Rank Adaptation (LoRA)~\cite{hu2022lora}, demonstrating strong performance across video, audio, depth, and infrared modalities. LanguageBind outperforms ImageBind in infrared, depth, and audio classification tasks.

Although LanguageBind adopts text as the anchor modality—similar to our approach—it also relies primarily on Vision Transformers (ViT) as the backbone architecture across modalities. This design choice, mirroring a key limitation of ImageBind, reduces flexibility and may hinder the integration of specialized pretrained encoders that could better capture modality-specific characteristics.

Vision-Audio-Language Omni-peRception (\textbf{VA\-LOR})~\cite{liu2024valor} advances multimodal research by integrating vision, audio, and language within a tri-modal framework. It introduces two pretext tasks: Multimodal Grouping Alignment for fine-grained modality alignment and Multimodal Grouping Captioning for text generation based on different modality combinations. VALOR established robust alignment between modalities and support tasks such as retrieval, captioning, and question-answering.

Vision-Audio-Subtitle-Text omni-modality foundation model (\textbf{VAST})~\cite{chen2023vast} expands multimodal learning by integrating vision, audio, subtitles, and text into a unified framework. By integrating subtitles and auxiliary modalities, VAST addressed the limitations of prior work, which often overlooked the role of additional information streams in video understanding, thereby highlighting the importance of datasets and models that leverage multiple complementary sources of information.

Unlike anchor-based approaches such as ImageBind and LanguageBind, VAST employs a joint training strategy that simultaneously optimizes across all modalities (e.g., video, audio, and text). Although this enables direct cross-modal interaction and potentially richer representations, it significantly increases computational demands during training, requiring substantial infrastructure and resource allocation. Moreover, the lack of a designated anchor modality may complicate scalability and alignment consistency when incorporating additional or low-resource modalities.

Multilingual Multimodal Pretraining (\textbf{MLMM})~\cite{zhang2023multimodal} advances multilingual multimodal pretraining by addressing the predominance of English in existing models. It combines pretraining-based and generalization-based approaches. For pretraining, MLMM leverages large-scale multilingual image-text datasets with texts translated into multiple languages. It employs four key pretraining objectives: Image-Text Matching for coarse-grained alignment, Masked Language Modeling for fine-grained cross-modal understanding, Masked Region Feature Regression, and Masked Region Classification for vision-language alignment. 

MLMM demonstrates strong cross-lingual transfer, particularly when fine-tuned with languages from the same language family. MLMM also explores generalization-based approaches through multilingual knowledge distillation and multilingual acquisition as resource-efficient alternatives, achieving state-of-the-art performance across multilingual vision-language tasks while maintaining deployment flexibility.

Connecting Multi-modal Contrastive Representations (\textbf{C-MCR})~\cite{wang2023connecting} proposes a method for aligning modalities using an overlapping modality as an anchor. The framework introduces two processes: inter-MCR, which connects overlapping modalities, and intra-MCR, which preserves relationships among non-overlapping ones. However, this approach requires the availability of multiple fully trained bimodal models to integrate each new modality. As a result, extending the system requires not only adding new encoders but also developing a comprehensive alignment procedure for each new modality, thereby increasing both implementation complexity and computational cost.

As evidenced by the literature, the proliferation of both multimodal and multilingual models is undeniable. However, models that truly excel at simultaneously integrating multimodality and multilinguality remain relatively scarce. Moreover, the few existing models that attempt this integration are often plagued by substantial training costs, long training times, and significant computational demands. Table~\ref{tab:related-work} overviews the related models' main features.

\begin{table}[!htb]
\caption{Comparison of CACARA with representative multimodal and multilingual models. ``Multilingual support'' refers to the model’s ability to operate in multiple languages. ``Free joint learning'' indicates models that do not require fully aligned data across three or more modalities. ``Pretrained encoders'' denotes the reuse of pretrained encoders without full retraining. ``Reduced training cost'' refers to models designed to reduce resource usage. ``Scalable to new modalities'' refers to models that allow adding new modalities without retraining the full architecture.  
}
\resizebox{0.85\columnwidth}{!}{%
{\fontsize{11pt}{12pt}\selectfont
\begin{tabular}{lccccc}
\hline
\multirow{2}{*}{Model} & Multilingual & Free joint & Pretrained & Reduced & Scalable to \vspace{-0.1cm}\\ 
 & support & learning & encoders & training cost & new modalities\\
\hline
CLIP                       & \xmark                                                          & \cmark                                                         & \xmark                                                         & \xmark                                                                & \xmark                                                           \\
CAPIVARA                   & \xmark                                                    & \cmark                                                         & \cmark                                                         &    \cmark                                                             &       \xmark                                                     \\
ImageBind                  & \xmark                                                          & \cmark                                                         & \xmark                                                         & \xmark                                                                & \xmark                                                           \\
LanguageBind               & \xmark                                                          & \cmark                                                         & \xmark                                                         & \xmark                                                                & \xmark                                                           \\
VALOR                      & \xmark                                                          & \xmark                                                         & \xmark                                                         & \xmark                                                                & \xmark                                                           \\
VAST                       & \xmark                                                          & \xmark                                                         & \xmark                                                         & \xmark                                                                & \xmark                                                           \\
MLMM                       & \cmark                                                          & \xmark                                                         & \cmark                                                         & \xmark                                                                & \xmark                                                           \\
C-MCR                      & \xmark                                                          & \cmark                                                         & \cmark                                                         & \xmark                                                                & \xmark                                                           \\
CACARA                    & \cmark                                                & \cmark                                                         & \cmark                                                         & \cmark                                                                & \cmark                                                           \\ \hline
\end{tabular}%
}
}
\label{tab:related-work}
\end{table}





\textbf{CACARA} directly addresses these limitations by introducing a novel and efficient approach based on an implicit learning strategy. This strategy not only facilitates seamless transfer learning across diverse modalities but also demonstrates the effective transfer of extensive multilingual capabilities from the anchor text encoder to newly introduced modalities.

By performing alignment using exclusively English data for the new modality (audio, in our current work), CACARA allows this modality to inherit support for over 100 languages from the text anchor, without requiring direct multilingual audio training data or retraining of the anchor itself. This highlights an efficient pathway for expanding multilinguality. Regarding joint training, the primary constraint lies in the scarcity of large-scale datasets offering three or more simultaneously aligned modalities, a limitation acknowledged in related works (e.g., ImageBind, LanguageBind). The proposed sequential alignment strategy addresses this practical limitation while also providing computational benefits.

\section{Methodology}
\label{sec:method}

This section presents the CACARA model's overall framework, covering its architectural design, training and evaluation datasets, and selected hyperparameters. We then detail the training workflow, highlighting the emergent alignment strategy responsible for its multimodal and multilingual capabilities.

\subsection{CACARA Model}

\begin{figure*}[!htb]
\includegraphics[width=1\linewidth]{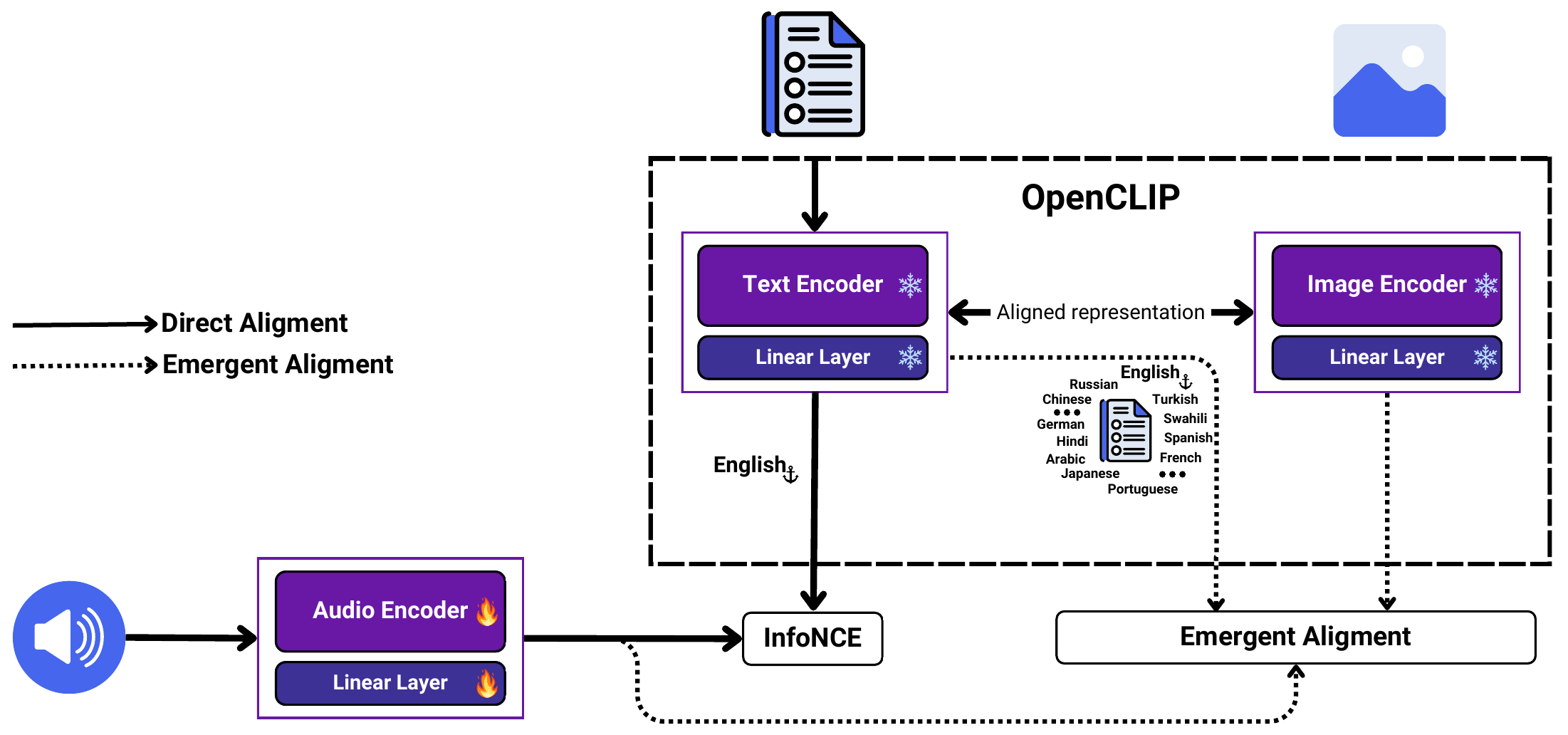}
    \caption{Flow of adding a new modality (Audio) to the text-image bimodal model. Linguistic expansion and alignment of these languages for the new modality. During audio model training, alignment is performed only with the anchor encoder (textual). At this stage, the model does not process any image information. In addition to the 12 highlighted languages selected for translation and evaluation, the model supports over 100 languages in total.}
    \label{fig:training}
\end{figure*}

The CACARA model integrates multimodal and multilingual learning through three encoders: image, text, and audio (Figure~\ref{fig:training}). The image encoder is based on a Vision Transformer (ViT)~\cite{alexey2021vit}, while the text encoder utilizes XLM-RoBERTa (base version)~\cite{conneau-etal-2020-unsupervised}. The audio encoder, built upon BEATs~\cite{chen2023beats}, is incorporated via emergent alignment learning. The image and text encoders are initialized with pretrained OpenCLIP~\cite{openclip_2021} encoders.

Training is performed using contrastive learning with the InfoNCE~\cite{oord2018infonce} loss function to align the audio and text encoders. 
\begin{equation}
\label{equation:InfoNCE}
    \mathcal{L}_{\text{N}} = - \mathbb{E}_{X} \left[ \log \frac{f_k(x_{t+k}, c_t)}{\sum\limits_{x_j \in X} f_k(x_j, c_t)} \right],
\end{equation}
\noindent where $X={x_1,\ldots,x_N}$ is a set of $N$ random samples from different modalities—$x$ denotes the target and $c$ the context—comprising one positive sample drawn from the conditional distribution $p(x_{t+k}\mid c_t)$ (with $x_{t+k}$ a future observation and $c_t$ the current context) and $N-1$ negative samples unrelated to $c_t$. Here, $f_k$ denotes the similarity function.

We keep the image and text models frozen to preserve the high-quality image-text alignment from OpenCLIP pretraining, optimizing only the audio encoder's alignment within the shared feature space. A key aspect of CACARA's training procedure is that, unlike LiT, which adapts a text encoder for downstream tasks while keeping the image encoder fixed, CACARA maintains the pretrained image-text representation and aligns the new audio encoder to this existing joint~space. 

\subsection{Multimodal and Multilingual Emergent Alignment}

CACARA training leverages the emergent learning capacity of pretrained models. Since the image and text encoders are pre-aligned in a shared feature space due to their pretraining, training the newly incorporated audio model against the text encoder implicitly aligns it with the shared space of all three. This emergent alignment enables the new audio model to describe untrained image modality features without training.

The text modality serves as the primary anchor for emergent learning in this architecture. As depicted in Figure~\ref{fig:training}, solid lines indicate direct training pairings (audio-text data presented to the model), while dotted lines indicate emergent learning, the emergent relationship between audio and image.

Due to the text encoder's role as an anchor and the freezing of pretrained text and image encoders, the text encoder remains fixed, synchronizing with additional modalities. This eliminates the need for explicit multilingual training, as XLM-RoBERTa's inherent multilingual capabilities extend to additional modalities through emergent alignment. Integrating a new modality synchronizes it with the multilingual features of the textual model.

This strategy enables non-multilingual models, such as BEATs, to acquire linguistic capabilities from the text model. As illustrated in Figure~\ref{fig:training}, multiple languages align with other modalities via this emergent synchronization, significantly reducing training costs and time. Unlike conventional approaches that train models on each language separately, CACARA requires training only in English, leveraging its higher data availability and model quality. This contrasts with existing literature, which often incurs multiplicative computational costs by training on multilingual datasets individually.

Although CACARA's text encoder supports over $100$ languages, we selected $12$ languages for evaluation: English, Portuguese, Spanish, French, German, Chinese, Japanese, Russian, Turkish, Hindi, Arabic, and Swahili. Despite an extensive search, we did not identify any publicly available multilingual, human-annotated datasets for bimodal audio-text tasks, nor any multilingual datasets containing more than two modalities, including audio. Consequently, we adopted an evaluation methodology based on automatically translated data. Specifically, in the absence of multilingual test sets, we translated the original English test data into the target languages using Google Translate. 

\subsection{Training Pipeline}

\begin{figure}[!htb] \includegraphics[width=0.9\linewidth]{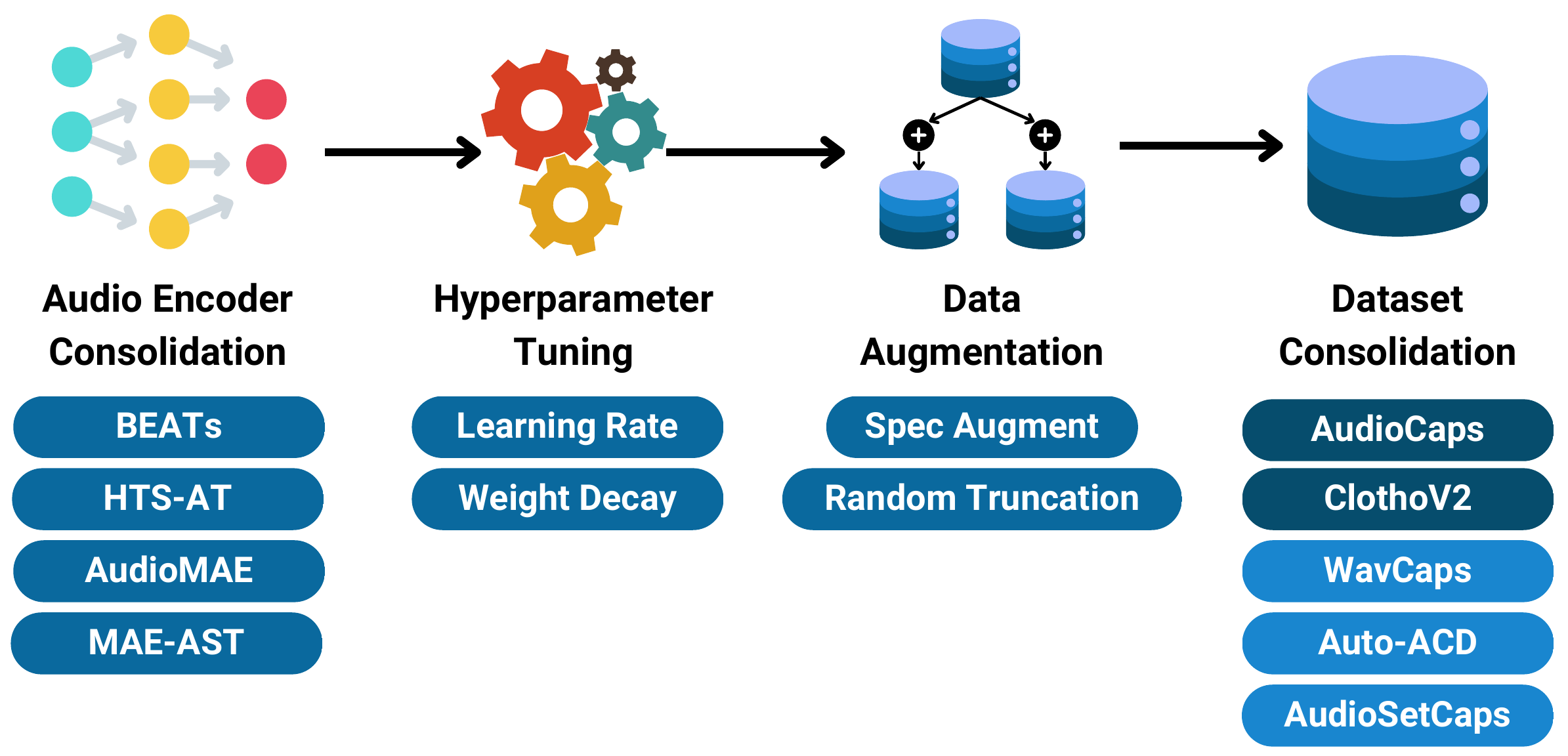}
    \caption{The CACARA model's training steps are divided into four main stages: consolidation of the newly added encoder, hyperparameter tuning, data augmentation, and consolidation of the training datasets.}
    \label{fig:experimentsv2}
\end{figure}

We developed the final CACARA model via a four-stage training pipeline (Figure~\ref{fig:experimentsv2}). The first stage involved a comparative evaluation of different audio encoders to select the optimal one for our framework and target tasks. For that, we selected four encoders: BEATs~\cite{chen2023beats}, HTS-AT~\cite{chen2022hts}, AudioMAE~\cite{huang2022masked}, and MAE-AST~\cite{baade22_interspeech}. We chose these models based on their state-of-the-art performance in sound event detection and audio tagging, and their relatively recent introduction to the field, as established in the existing literature. Section~\ref{sec:audio_encoders} provides a detailed analysis of the models built with these encoder combinations.

The second stage focused on identifying optimal hyperparameters. While many parameters were investigated, the learning rate and weight decay substantially influenced training and model performance. We conducted a systematic search to determine values that ensured a fair comparison across all encoders. Appendix \hyperlink{appendixA}{A} describes the best combination of hyperparameters.


The third stage incorporated data augmentation to enhance the model's robustness. We employed two data augmentation strategies operating on the audio modality: Random Truncation (RT) and SpecAugment. Our choice was guided by their proven effectiveness in unimodal audio tasks~\cite{kong2020panns, gong2021psla, ferreira2023automatic} and multimodal scenarios~\cite{CLAP2023}. Specifically, Random Truncation involves randomly segmenting audio inputs to a fixed length, while SpecAugment masks blocks of information in the time and frequency domains. Section~\ref{sec:augmentation} provides an analysis of configurations that use these augmentation methods, and Appendix \hyperlink{appendixB}{B} details these techniques.


`The fourth stage involved a deliberate selection of datasets, focusing on quantity, diversity, multimodal decoupling, and label quality, to evaluate various combinations to identify the subset yielding optimal performance. We chose five datasets: AudioCaps~\cite{kim-etal-2019-audiocaps}, ClothoV2~\cite{drossos2020clotho}, WavCaps~\cite{mei2024wavcaps}, Auto-ACD~\cite{sun2024auto}, and AudioSetCaps~\cite{bai2024audiosetcaps}. Details about these datasets are presented in Appendix \hyperlink{appendixC}{C}.


AudioCaps and ClothoV2, both featuring human-annotated captions, served as gold standards and were included in all dataset configurations. We augmented this base with various combinations of machine-annotated datasets to assess their performance impact, with specific configurations and results detailed in Section~\ref{subsec:multimodal-training}. We evaluated the model on two core tasks: retrieval (using AudioCaps and ClothoV2) and classification (using ESC-50~\cite{piczak2015dataset} and UrbanSounds8K~\cite{diment_2017_401395}). Retrieval performance was measured using R-precision at ranks 1, 5, and 10 (R@1, R@5, R@10) and mean average recall (R@Avg), while classification performance was measured using mean accuracy.

\section{Experiments and Results}
\label{sec:experiments}

Our analysis of CACARA focuses on four primary characteristics: (1) its multimodal capabilities, (2)~its multilingual performance, (3) the efficiency and scaling of its underlying resources, and (4)~expanded resources. We discussed the first two characteristics and their results in Sections~\ref{subsec:multimodal-training} and~\ref{subsec:multilingual-analysis}. To illustrate the computational efficiency of our approach compared to training a fully tri-modal model, we report the resources our model requires in Section~\ref{subsec:compute-resources}. In addition, to compare and understand the model's capabilities regarding resources and performance, we have conducted extended tests that demonstrate how the model behaves in scenarios with greater computational capacity, as detailed in Section~\ref{subsec:extended-resources}. Intermediate training results and ablation studies on model components are presented in Section~\ref{sec:audio_encoders}, Appendices \hyperlink{appendixA}{A} and \hyperlink{appendixB}{B}, as well as extended results with other sets and combinations of data. Qualitative visualizations are provided in Section~\ref{sec:appendix_qualitative_analysis}.

For the results presented in this section, we used the training datasets AudioSetCaps (ASC), Auto-ACD (AA), WavCaps (WC), AudioCaps (AC), and ClothoV2 (C). For the CACARA model, different combinations of datasets have been evaluated, yielding varying results depending on the task and on distributions similar to those of the training data. We also applied data filtering based on CLIP similarity, where the filtering percentage x\% is specified as f~0.x. 

\subsection{Audio Encoder Consolidation}
\label{sec:audio_encoders}

As the first step in the CACARA model training pipeline, we focused on selecting the audio encoder to be incorporated into the multimodal model. The results obtained for each of these models are listed in Table~\ref{tab:models-ablations-retrieval}, for the retrieval task, and for classification in Table~\ref{tab:models-ablations-classification}. From this set of experiments, we observed that the BEATs encoder consistently achieved the best performance, with variations only across training configurations. For audio--text retrieval, on the AudioCaps dataset, the $\text{BEATs}_{ASC/AA/W}$ yielded improvements of up to $4$ pp, while on ClothoV2, the $\text{BEATs}_{wc/f~0.2}$ outperformed others by up to $10$ pp. In the classification tasks, using ESC-50 and UrbanSounds8K, the $\text{BEATs}_{wc/f~0.2}$ again achieved the best results across both datasets. Based on these findings, we selected the BEATs model for subsequent experiments, as it demonstrated superior performance after integration with the image and text encoders.

\begin{table}[t]
\caption{Ablation results for the different audio encoders tested in the audio encoder consolidation phase for the audio-to-text and text-to-audio retrieval tasks, on the AudioCaps and ClothoV2~datasets.}
\resizebox{0.9\columnwidth}{!}{%
\begin{tabular}{lcccccccc}
\hline
                          & \multicolumn{4}{c|}{Audio-to-Text}                 & \multicolumn{4}{c}{Text-to-Audio} \\ \cline{2-9} 
                          & \multicolumn{8}{c}{AudioCaps}                                  \\ \cline{2-9} 
\multicolumn{1}{c}{Model} & R@1   & R@5   & R@10  & \multicolumn{1}{c|}{R@Avg} & R@1    & R@5    & R@10   & R@Avg  \\ \hline
AudioMAE                  & 27.29 & 59.08 & 74.74 & \multicolumn{1}{c|}{53.70} & 5.47   & 22.41  & 34.11  & 20.66  \\
HTSAT                     & 10.21 & 30.55 & 43.22 & \multicolumn{1}{c|}{27.99} & 1.69   & 7.63   & 14.31  & 7.87   \\
MAE\_AST                   & 26.84 & 61.08 & 76.04 & \multicolumn{1}{c|}{54.65} & 5.49   & 22.34  & 34.02  & 20.61  \\
$\text{BEATs}_{wc/f~0.2} $         & 30.44 & 64.99 & 78.22 & \multicolumn{1}{c|}{57.89} & 7.40   & 28.08  & 42.56  & 26.01  \\
$\text{BEATs}_{ASC/AA/W}$                     & 34.69 & 68.75 & 81.57 & \multicolumn{1}{c|}{61.67} & 7.24   & 27.85  & 41.17  & 25.42  \\ \hline
                          & \multicolumn{8}{c}{ClothoV2}                                   \\ \cline{2-9} 
AudioMAE                  & 10.55 & 27.85 & 39.46 & \multicolumn{1}{c|}{25.95} & 2.16   & 8.63   & 14.30  & 8.36   \\
HTSAT                     & 4.19  & 14.30 & 22.12 & \multicolumn{1}{c|}{13.54} & 0.61   & 2.64   & 4.90   & 2.72   \\
MAE\_AST                   & 9.91  & 26.81 & 37.89 & \multicolumn{1}{c|}{24.87} & 2.22   & 9.21   & 15.06  & 8.83   \\
$\text{BEATs}_{wc/f~0.2}$          & 19.94 & 47.42 & 61.33 & \multicolumn{1}{c|}{42.90} & 4.42   & 17.04  & 27.51  & 16.32  \\
$\text{BEATs}_{ASC/AA/W}$          & 9.45  & 25.70 & 36.13 & \multicolumn{1}{c|}{23.76} & 1.56   & 6.59   & 10.94  & 6.36   \\
\hline
\end{tabular}%
}
\label{tab:models-ablations-retrieval}
\end{table}

\begin{table}[t]
\caption{Comparison between the different encoders tested in the audio encoder consolidation phase, for classification task on the ESC-50 and UrbanSounds8K datasets.}
\label{tab:models-ablations-classification}
\resizebox{0.5\columnwidth}{!}{%
\begin{tabular}{lcc}
\hline
\multicolumn{1}{c}{}              & ESC-50    & UrbanSounds8K    \\ \cline{2-3} 
\multicolumn{1}{c}{Model} & \multicolumn{2}{c}{Accuracy (\%)} \\
\hline
AudioMAE                          & 67.80      & 65.34            \\
HTSAT                             & 34.20      & 48.16            \\
MAE\_AST                          & 67.25      & 66.92            \\
$\text{BEATs}_{wc/f~0.2}$         & 93.25      & 77.87            \\
$\text{BEATs}_{ASC/AA/W}$         & 90.90      & 75.62            \\
\hline
\end{tabular}%
}
\end{table}

\subsection{Data Augmentation}
\label{sec:augmentation}

We conducted a comprehensive ablation study to evaluate the effectiveness of the selected techniques, Random Truncation and SpecAugment, both individually and in combination. The results for retrieval and classification tasks are presented in Tables~\ref{tab:augumentation-retrieval} and~\ref{tab:augumentation-classification}, respectively.

For the retrieval tasks detailed in Table~\ref{tab:augumentation-retrieval}, the combination of both augmentation methods proved most effective. On the AudioCaps dataset, the combined SpecAugment + RT strategy yielded the best performance in audio-to-text retrieval, improving the R@Avg score to $66.28$. Interestingly, applying either SpecAugment or RT individually resulted in a slight performance degradation, highlighting a synergistic benefit of using both together. For text-to-audio retrieval on AudioCaps and for all retrieval tasks on ClothoV2, every augmentation configuration outperformed the baseline model.

The benefits of data augmentation were even more pronounced for the classification tasks, as shown in Table~\ref{tab:augumentation-classification}. On the ESC-50 dataset, combining SpecAugment and RT resulted in a substantial performance gain, increasing the classification accuracy by $6.75$ pp, from $76.15\%$ to $82.90\%$. A similar trend was observed on the UrbanSounds8K dataset, where the combined approach improved accuracy by $1.55$ pp over the baseline.

Given these consistent and significant improvements, we adopted the combined SpecAugment + RT strategy for subsequent experiments.

\begin{table}[t]
\caption{Ablation results for the different datasets tested in the training dataset selection phase, for the audio-to-text and text-to-audio retrieval tasks, on the AudioCaps and ClothoV2 datasets.}
\resizebox{\columnwidth}{!}{%
\begin{tabular}{lcccccccc}
\hline
                & \multicolumn{4}{c|}{Audio-to-Text}                 & \multicolumn{4}{c}{Text-to-Audio} \\ \cline{2-9} 
                & \multicolumn{8}{c}{AudioCaps}                                       \\
\cline{2-9}
\multicolumn{1}{c}{Model} & R@1   & R@5   & R@10  & \multicolumn{1}{c|}{R@Avg} & R@1  & R@5   & R@10  & R@Avg \\ \hline
BEATs + No-Augument       & 31.09 & 66.07 & 79.71 & \multicolumn{1}{c|}{66.07} & 6.25 & 24.86 & 38.11 & 24.86 \\
BEATs + SpecAug & 31.05 & 65.60 & 79.55 & \multicolumn{1}{c|}{65.60} & 6.59   & 25.53  & 38.45  & 25.53  \\
BEATs + RT      & 30.33 & 65.15 & 80.09 & \multicolumn{1}{c|}{65.15} & 6.21   & 25.51  & 38.13  & 25.51  \\
BEATs + SpecAug + RT      & 32.01 & 66.28 & 79.80 & \multicolumn{1}{c|}{66.28} & 6.32 & 25.40 & 39.46 & 25.40 \\ \hline
                & \multicolumn{8}{c}{ClothoV2}                                        \\ \cline{2-9} 
BEATs + No-Augument       & 10.12 & 28.98 & 41.63 & \multicolumn{1}{c|}{28.98} & 2.51 & 9.68  & 15.90 & 9.68  \\
BEATs + SpecAug & 11.25 & 30.18 & 41.93 & \multicolumn{1}{c|}{30.18} & 2.76   & 10.32  & 17.03  & 10.32  \\
BEATs + RT      & 10.41 & 29.00 & 41.17 & \multicolumn{1}{c|}{29.00} & 2.49   & 10.30  & 16.59  & 10.30  \\
BEATs + SpecAug + RT      & 11.12 & 29.89 & 41.97 & \multicolumn{1}{c|}{29.89} & 2.51 & 9.97  & 15.87 & 9.97  \\ 
\hline
\end{tabular}%
}
\label{tab:augumentation-retrieval}
\end{table}

\begin{table}[t]
\caption{Ablation results for the classification task between the different datasets tested in the training dataset selection phase, on the ESC-50 and ClothoV2 datasets.}
\label{tab:augumentation-classification}
\resizebox{0.6\columnwidth}{!}{%
\begin{tabular}{lcc}
\hline
\multicolumn{1}{c}{}              & ESC-50    & UrbanSounds8K    \\ \cline{2-3}
\multicolumn{1}{c}{Augmentation} & \multicolumn{2}{c}{Accuracy (\%)} \\
\hline
BEATs + No-Augument               & 76.15     & 62.31            \\
BEATs + SpecAug                   & 79.05     & 62.76            \\
BEATs + RT                        & 80.95     & 60.49            \\
BEATs + SpecAug + RT              & 82.90     & 63.86            \\
\hline
\end{tabular}%
}
\end{table}

\subsection{Multimodal Evaluation}
\label{subsec:multimodal-training}

We compare CACARA with established bimodal and multimodal models. While most prior work has focused on bimodal architectures, these models inherently lack the flexibility to handle inter-domain scenarios and emergent learning, where modalities independently acquire new conceptual representations. In contrast, CACARA is designed to leverage multimodality, expanding its applicability and enhancing adaptability across diverse tasks.

We selected three representative bimodal models for comparison: CLAP (Microsoft)~\cite{CLAP2023}, CLAP (LAION)~\cite{laionclap2023}, and WavCaps Model~\cite{mei2023wavcaps}. These models, all focusing on audio-text modalities, provide a relevant benchmark for evaluating the audio-centric capabilities introduced in CACARA. This comparison aims to leverage these bimodal models' high degree of alignment and reported performance as a reference point for achievable results within a constrained modality space. For this reason, bimodal models cannot be directly compared to multimodal models. Thus, the tables highlight the best results specifically for multimodal models.

For the multimodal comparison, we selected ImageBind, VAST, and LanguageBind models, which, to the best of our knowledge, represent the state of the art in multimodal learning. A direct comparison with MLMM, the sole identified work in our review to integrate both multimodality and multilinguality, was not feasible due to the lack of publicly accessible code and implementation details, hindering reproducibility. Thus, our comparison remains direct and comprehensive, focusing on models that effectively align multiple modalities.


\begin{figure}[t]
\includegraphics[width=1\linewidth]{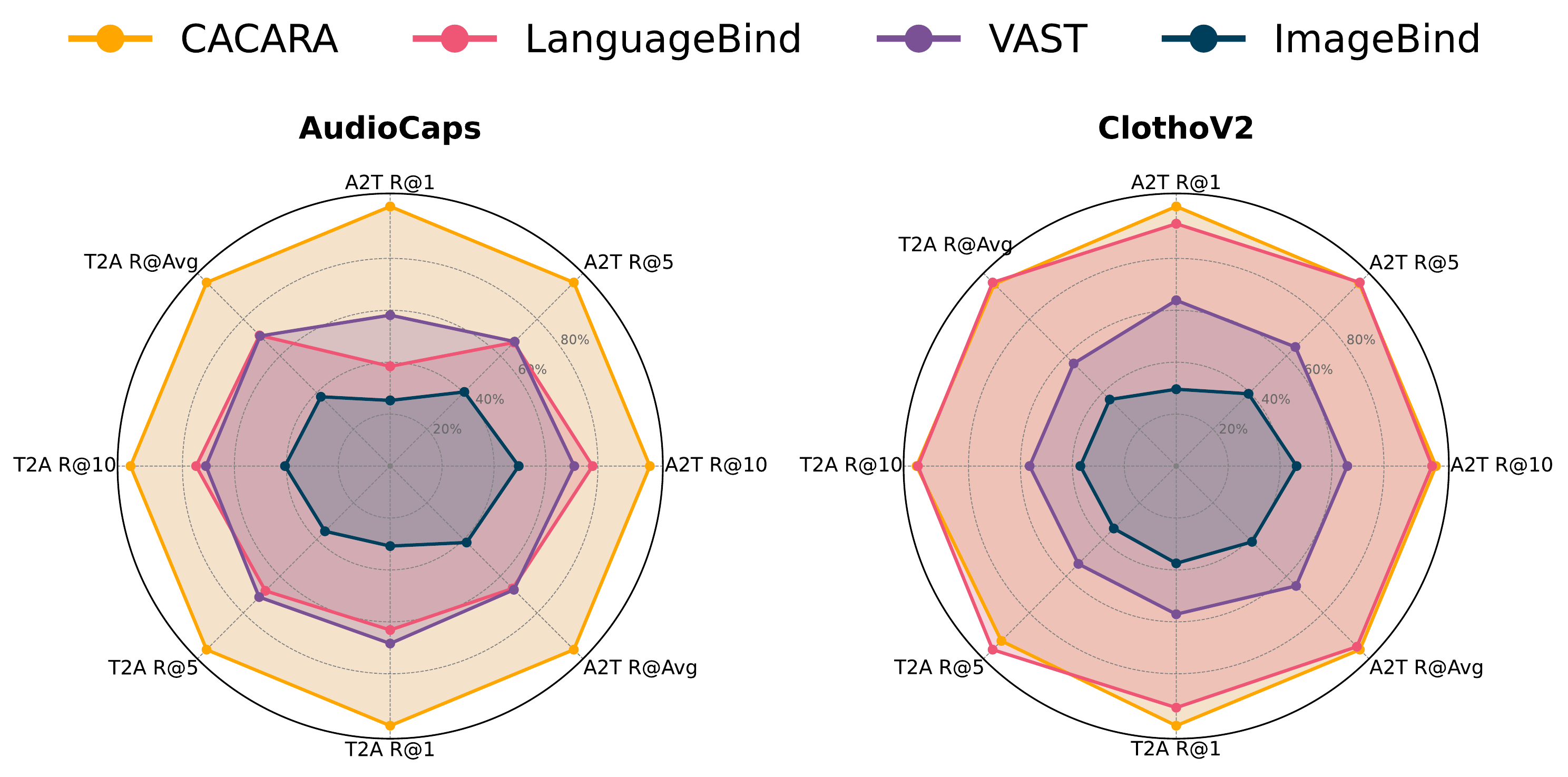}
    \caption{Performance comparison of ImageBind, VAST, LanguageBind, and CACARA on Audio-to-Text and Text-to-Audio retrieval using AudioCaps and ClothoV2 datasets. CACARA variants: $\text{CACARA}_{ASC/AA/WC/AC/C/f~0.2}$ (AudioCaps) and $\text{CACARA}_{WC/AC/C/f~0.2}$ (ClothoV2). Metrics show normalized R@1, R@5, R@10, and R@Avg scaled between min-max values.}
    \label{fig:radial-multimodal-training}
\end{figure}

Based on the results summarized in Table~\ref{tab:multimodal-training}, Figure~\ref{fig:radial-multimodal-training} presents a normalized radar chart representation, enabling visual comparison of model performance (R@1, R@5, R@10, R@Avg) across datasets. The models compared are ImageBind, LanguageBind, VAST, and CACARA.

Table~\ref{tab:multimodal-training} shows results in cross-modal retrieval tasks. Audio-to-text retrieval refers to the task of retrieving the most relevant textual description given an audio query, whereas text-to-audio retrieval involves retrieving the corresponding audio segment given a textual query. For Audio\-Caps, CACARA, trained on all datasets with a $0.2$ filtering threshold, achieved the highest R@1 ($33.98\%$) among multimodal models, surpassing the best existing multimodal models by $14.23$ pp while keeping competitive performance against bimodal models. For ClothoV2, the best-performing model was CACARA trained with WavCaps, with R@1 of $17.26\%$.


\begin{table*}[!htb]
\caption{Multimodal results, divided into two datasets: AudioCaps and ClothoV2, and two retrieval tasks: audio-to-text and text-to-audio. The evaluated models are categorized as BM (Bimodal Models), MM (Multimodal Models), and CACARA (CACARA Multimodal Models). \textbf{Bolded} results indicate the best performance among multimodal models, while \underline{underlined} results represent the second-best values for the same set.}
\resizebox{1\textwidth}{!}{%
\begin{tabular}{llclllllllllllllllllllllll}
\hline
\multicolumn{2}{c}{\multirow{3}{*}{Model}} &
  \multicolumn{12}{c}{Audio-to-Text} &
  \multicolumn{12}{c}{Text-to-Audio} \\ \cline{3-26} 
\multicolumn{2}{c}{} &
  \multicolumn{3}{c}{R@1} &
  \multicolumn{3}{c}{R@5} &
  \multicolumn{3}{c}{R@10} &
  \multicolumn{3}{c}{R@Avg} &
  \multicolumn{3}{c}{R@1} &
  \multicolumn{3}{c}{R@5} &
  \multicolumn{3}{c}{R@10} &
  \multicolumn{3}{c}{R@Avg} \\ \hline
\multicolumn{2}{l}{} &
  \multicolumn{24}{c}{AudioCaps} \\ \cline{3-26} 
\multirow{3}{*}{\rotatebox[origin=c]{90}{BM}} &
  CLAP (Microsoft) &
  \multicolumn{3}{c}{15.75} &
  \multicolumn{3}{c}{44.7} &
  \multicolumn{3}{c}{61.62} &
  \multicolumn{3}{c}{40.69} &
  \multicolumn{3}{c}{6.32} &
  \multicolumn{3}{c}{24.9} &
  \multicolumn{3}{c}{38.11} &
  \multicolumn{3}{c}{23.11} \\
 &
  CLAP (LAION) &
  \multicolumn{3}{c}{34.58} &
  \multicolumn{3}{c}{70.8} &
  \multicolumn{3}{c}{83.69} &
  \multicolumn{3}{c}{63.02} &
  \multicolumn{3}{c}{9.31} &
  \multicolumn{3}{c}{35.52} &
  \multicolumn{3}{c}{51.68} &
  \multicolumn{3}{c}{32.17} \\
 &
  WavCaps &
  \multicolumn{3}{c}{38.70} &
  \multicolumn{3}{c}{73.32} &
  \multicolumn{3}{c}{86.05} &
  \multicolumn{3}{c}{66.02} &
  \multicolumn{3}{c}{10.57} &
  \multicolumn{3}{c}{38.38} &
  \multicolumn{3}{c}{53.21} &
  \multicolumn{3}{c}{34.05} \\ \hline
\multirow{3}{*}{\rotatebox[origin=c]{90}{MM}} &
  ImageBind &
  \multicolumn{3}{c}{8.59} &
  \multicolumn{3}{c}{27.58} &
  \multicolumn{3}{c}{40.49} &
  \multicolumn{3}{c}{25.55} &
  \multicolumn{3}{c}{2.25} &
  \multicolumn{3}{c}{9.90} &
  \multicolumn{3}{c}{16.69} &
  \multicolumn{3}{c}{9.61} \\
 &
  VAST &
  \multicolumn{3}{c}{19.75} &
  \multicolumn{3}{c}{46.30} &
  \multicolumn{3}{c}{57.98} &
  \multicolumn{3}{c}{41.34} &
  \multicolumn{3}{c}{4.99} &
  \multicolumn{3}{c}{19.89} &
  \multicolumn{3}{c}{29.25} &
  \multicolumn{3}{c}{18.04} \\
 &
  LanguageBind &
  \multicolumn{3}{c}{13.05} &
  \multicolumn{3}{c}{45.96} &
  \multicolumn{3}{c}{63.76} &
  \multicolumn{3}{c}{40.92} &
  \multicolumn{3}{c}{4.61} &
  \multicolumn{3}{c}{18.94} &
  \multicolumn{3}{c}{30.80} &
  \multicolumn{3}{c}{18.12} \\ \hline
\multirow{7}{*}{\rotatebox[origin=c]{90}{CACARA}} &
  $\text{CACARA}_{AA/WC/AC/C}$ &
  \multicolumn{3}{c}{31.03 \scriptsize{± 0.51}} &
  \multicolumn{3}{c}{64.57 \scriptsize{± 0.74}} &
  \multicolumn{3}{c}{79.34 \scriptsize{± 0.53}} &
  \multicolumn{3}{c}{58.31 \scriptsize{± 0.43}} &
  \multicolumn{3}{c}{6.76 \scriptsize{± 0.45}} &
  \multicolumn{3}{c}{25.52 \scriptsize{± 1.65}} &
  \multicolumn{3}{c}{38.61 \scriptsize{± 1.30}} &
  \multicolumn{3}{c}{23.63 \scriptsize{± 1.13}} \\
 &
  $\text{CACARA}_{WC/AC/C}$ &
  \multicolumn{3}{c}{30.08 \scriptsize{± 0.31}} &
  \multicolumn{3}{c}{64.95 \scriptsize{± 0.42}} &
  \multicolumn{3}{c}{78.71 \scriptsize{± 0.75}} &
  \multicolumn{3}{c}{57.91 \scriptsize{± 0.41}} &
  \multicolumn{3}{c}{\textbf{7.57} \scriptsize{±0.16}} &
  \multicolumn{3}{c}{{\ul 28.02} \scriptsize{± 0.23}} &
  \multicolumn{3}{c}{{\ul 42.43} \scriptsize{± 0.14}} &
  \multicolumn{3}{c}{{\ul 26.00} \scriptsize{± 0.09}} \\
 &
  $\text{CACARA}_{WC/AC/C/f~0.1}$ &
  \multicolumn{3}{c}{30.96 \scriptsize{± 0.05}} &
  \multicolumn{3}{c}{64.78 \scriptsize{± 0.91}} &
  \multicolumn{3}{c}{78.54 \scriptsize{± 0.35}} &
  \multicolumn{3}{c}{58.10 \scriptsize{± 0.41}} &
  \multicolumn{3}{c}{7.37 \scriptsize{± 0.20}} &
  \multicolumn{3}{c}{27.88 \scriptsize{± 0.26}} &
  \multicolumn{3}{c}{41.72 \scriptsize{± 0.61}} &
  \multicolumn{3}{c}{25.66 \scriptsize{± 0.35}} \\
 &
  $\text{CACARA}_{WC/AC/C/f~0.2}$ &
  \multicolumn{3}{c}{31.03 \scriptsize{± 0.60}} &
  \multicolumn{3}{c}{65.32 \scriptsize{± 0.68}} &
  \multicolumn{3}{c}{78.80 \scriptsize{± 1.10}} &
  \multicolumn{3}{c}{58.38 \scriptsize{± 0.77}} &
  \multicolumn{3}{c}{{\ul 7.41} \scriptsize{±  0.09}} &
  \multicolumn{3}{c}{\textbf{28.40} \scriptsize{± 0.58}} &
  \multicolumn{3}{c}{\textbf{42.68}  \scriptsize{± 0.27}} &
  \multicolumn{3}{c}{\textbf{26.17}  \scriptsize{± 0.31}} \\
 &
  $\text{CACARA}_{ASC/AA/WC/AC/C}$ &
  \multicolumn{3}{c}{33.27 \scriptsize{± 0.33}} &
  \multicolumn{3}{c}{67.46 \scriptsize{± 0.33}} &
  \multicolumn{3}{c}{81.51 \scriptsize{± 0.30}} &
  \multicolumn{3}{c}{60.75 \scriptsize{± 0.29}} &
  \multicolumn{3}{c}{6.91 \scriptsize{± 0.15}} &
  \multicolumn{3}{c}{26.18 \scriptsize{± 0.18}} &
  \multicolumn{3}{c}{39.55 \scriptsize{± 0.15}} &
  \multicolumn{3}{c}{24.21 \scriptsize{± 0.06}} \\
 &
  $\text{CACARA}_{ASC/AA/WC/AC/C/f~0.1}$ &
  \multicolumn{3}{c}{{\ul 33.64} \scriptsize{± 0.24}} &
  \multicolumn{3}{c}{\textbf{68.40}  \scriptsize{± 0.24}} &
  \multicolumn{3}{c}{\textbf{81.84}  \scriptsize{± 0.17}} &
  \multicolumn{3}{c}{{\ul 61.29}  \scriptsize{± 0.23}} &
  \multicolumn{3}{c}{7.34  \scriptsize{± 0.10}} &
  \multicolumn{3}{c}{27.69  \scriptsize{± 0.32}} &
  \multicolumn{3}{c}{41.27  \scriptsize{± 0.29}} &
  \multicolumn{3}{c}{25.43  \scriptsize{± 0.05}} \\
 &
  $\text{CACARA}_{ASC/AA/WC/AC/C/f~0.2}$ &
  \multicolumn{3}{c}{\textbf{33.98} \scriptsize{± 0.64}} &
  \multicolumn{3}{c}{{\ul 68.30}  \scriptsize{± 0.64}} &
  \multicolumn{3}{c}{{\ul 81.81} \scriptsize{± 0.21}} &
  \multicolumn{3}{c}{\textbf{61.36} \scriptsize{± 0.26}} &
  \multicolumn{3}{c}{7.30  \scriptsize{± 0.15}} &
  \multicolumn{3}{c}{27.87  \scriptsize{± 0.32}} &
  \multicolumn{3}{c}{41.21  \scriptsize{± 0.29}} &
  \multicolumn{3}{c}{25.46  \scriptsize{± 0.16}} \\ \cline{2-26} 
 &
  \multicolumn{1}{c}{} &
  \multicolumn{24}{c}{ClothoV2} \\ \cline{3-26} 
\multirow{3}{*}{\rotatebox[origin=c]{90}{BM}} &
  CLAP (Microsoft) &
  \multicolumn{3}{c}{15.46} &
  \multicolumn{3}{c}{38.74} &
  \multicolumn{3}{c}{51.52} &
  \multicolumn{3}{c}{35.24} &
  \multicolumn{3}{c}{4.61} &
  \multicolumn{3}{c}{16.90} &
  \multicolumn{3}{c}{26.43} &
  \multicolumn{3}{c}{15.98} \\
 &
  CLAP (LAION) &
  \multicolumn{3}{c}{14.64} &
  \multicolumn{3}{c}{37.28} &
  \multicolumn{3}{c}{49.68} &
  \multicolumn{3}{c}{33.87} &
  \multicolumn{3}{c}{3.77} &
  \multicolumn{3}{c}{15.27} &
  \multicolumn{3}{c}{23.79} &
  \multicolumn{3}{c}{14.28} \\
 &
  WavCaps &
  \multicolumn{3}{c}{18.78} &
  \multicolumn{3}{c}{45.15} &
  \multicolumn{3}{c}{57.72} &
  \multicolumn{3}{c}{40.55} &
  \multicolumn{3}{c}{4.38} &
  \multicolumn{3}{c}{18.76} &
  \multicolumn{3}{c}{28.61} &
  \multicolumn{3}{c}{17.25} \\ \hline
\multirow{3}{*}{\rotatebox[origin=c]{90}{MM}} &
  ImageBind &
  \multicolumn{3}{c}{5.11} &
  \multicolumn{3}{c}{16.17} &
  \multicolumn{3}{c}{24.96} &
  \multicolumn{3}{c}{15.41} &
  \multicolumn{3}{c}{1.51} &
  \multicolumn{3}{c}{5.57} &
  \multicolumn{3}{c}{9.15} &
  \multicolumn{3}{c}{5.41} \\
 &
  VAST &
  \multicolumn{3}{c}{11.02} &
  \multicolumn{3}{c}{26.64} &
  \multicolumn{3}{c}{35.48} &
  \multicolumn{3}{c}{24.38} &
  \multicolumn{3}{c}{2.30} &
  \multicolumn{3}{c}{8.73} &
  \multicolumn{3}{c}{13.99} &
  \multicolumn{3}{c}{8.34} \\
 &
  LanguageBind &
  \multicolumn{3}{c}{{\ul 16.11}} &
  \multicolumn{3}{c}{\textbf{41.07}} &
  \multicolumn{3}{c}{{\ul 53.05}} &
  \multicolumn{3}{c}{{\ul 36.74}} &
  \multicolumn{3}{c}{3.75} &
  \multicolumn{3}{c}{\textbf{16.38}} &
  \multicolumn{3}{c}{{\ul 24.65}} &
  \multicolumn{3}{c}{\textbf{14.93}} \\ \hline
\multirow{7}{*}{\rotatebox[origin=c]{90}{CACARA}} &
  $\text{CACARA}_{AA/WC/AC/C}$ &
  \multicolumn{3}{c}{14.42 \scriptsize{± 0.88}} &
  \multicolumn{3}{c}{37.21 \scriptsize{± 1.74}} &
  \multicolumn{3}{c}{49.95 \scriptsize{± 1.60}} &
  \multicolumn{3}{c}{33.86 \scriptsize{± 1.40}} &
  \multicolumn{3}{c}{2.90 \scriptsize{± 0.32}} &
  \multicolumn{3}{c}{12.17 \scriptsize{± 1.49}} &
  \multicolumn{3}{c}{19.40 \scriptsize{± 1.75}} &
  \multicolumn{3}{c}{11.49 \scriptsize{± 1.18}} \\
 &
  $\text{CACARA}_{WC/AC/C}$ &
  \multicolumn{3}{c}{14.39 \scriptsize{± 0.19}} &
  \multicolumn{3}{c}{35.87 \scriptsize{± 0.21}} &
  \multicolumn{3}{c}{48.08 \scriptsize{± 0.32}} &
  \multicolumn{3}{c}{32.78 \scriptsize{± 0.14}} &
  \multicolumn{3}{c}{3.88 \scriptsize{± 0.13}} &
  \multicolumn{3}{c}{14.69 \scriptsize{± 0.41}} &
  \multicolumn{3}{c}{22.49 \scriptsize{± 0.12}} &
  \multicolumn{3}{c}{13.69 \scriptsize{± 0.18}} \\
 &
  $\text{CACARA}_{WC/AC/C/f~0.1}$ &
  \multicolumn{3}{c}{15.38 \scriptsize{± 0.49}}  &
  \multicolumn{3}{c}{37.40 \scriptsize{± 1.62}}  &
  \multicolumn{3}{c}{49.63 \scriptsize{± 2.13}}  &
  \multicolumn{3}{c}{34.13 \scriptsize{± 1.40}}  &
  \multicolumn{3}{c}{{\ul 3.92} \scriptsize{± 0.35}}  &
  \multicolumn{3}{c}{14.99 \scriptsize{± 0.41}}  &
  \multicolumn{3}{c}{23.10 \scriptsize{± 0.54}}  &
  \multicolumn{3}{c}{14.00 \scriptsize{± 0.42}}  \\
 &
  $\text{CACARA}_{WC/AC/C/f~0.2}$ &
  \multicolumn{3}{c}{\textbf{17.26} \scriptsize{± 2.33}} &
  \multicolumn{3}{c}{{\ul 40.91} \scriptsize{± 5.84}} &
  \multicolumn{3}{c}{\textbf{53.85} \scriptsize{± 6.74}} &
  \multicolumn{3}{c}{\textbf{37.34} \scriptsize{± 4.96}} &
  \multicolumn{3}{c}{\textbf{4.03} \scriptsize{± 0.36}} &
  \multicolumn{3}{c}{\textbf{15.60} \scriptsize{± 1.25}} &
  \multicolumn{3}{c}{\textbf{24.78} \scriptsize{± 2.36}} &
  \multicolumn{3}{c}{{\ul 14.80} \scriptsize{± 1.32}} \\
 &
  $\text{CACARA}_{ASC/AA/WC/AC/C}$ &
  \multicolumn{3}{c}{11.40 \scriptsize{± 2.17}} &
  \multicolumn{3}{c}{29.82 \scriptsize{± 6.17}} &
  \multicolumn{3}{c}{41.12 \scriptsize{± 7.79}} &
  \multicolumn{3}{c}{27.45 \scriptsize{± 5.37}} &
  \multicolumn{3}{c}{2.22 \scriptsize{± 0.61}} &
  \multicolumn{3}{c}{8.71 \scriptsize{± 2.41}} &
  \multicolumn{3}{c}{14.42 \scriptsize{± 3.66}} &
  \multicolumn{3}{c}{8.45 \scriptsize{± 2.22}} \\
 &
  $\text{CACARA}_{ASC/AA/WC/AC/C/f~0.1}$ &
  \multicolumn{3}{c}{13.04 \scriptsize{± 3.12}} &
  \multicolumn{3}{c}{33.53 \scriptsize{± 6.95}} &
  \multicolumn{3}{c}{46.00 \scriptsize{± 8.70}} &
  \multicolumn{3}{c}{30.86 \scriptsize{± 6.25}} &
  \multicolumn{3}{c}{2.56 \scriptsize{± 0.87}} &
  \multicolumn{3}{c}{9.84 \scriptsize{± 2.98}} &
  \multicolumn{3}{c}{16.15 \scriptsize{± 4.43}} &
  \multicolumn{3}{c}{9.52 \scriptsize{± 2.76}} \\
 &
  $\text{CACARA}_{ASC/AA/WC/AC/C/f~0.2}$ &
  \multicolumn{3}{c}{13.21 \scriptsize{± 3.16}} &
  \multicolumn{3}{c}{33.63 \scriptsize{± 6.91}} &
  \multicolumn{3}{c}{45.67 \scriptsize{± 8.51}} &   
  \multicolumn{3}{c}{30.84 \scriptsize{± 6.19}} &
  \multicolumn{3}{c}{2.59 \scriptsize{± 0.74}} &
  \multicolumn{3}{c}{10.07 \scriptsize{± 2.87}} &
  \multicolumn{3}{c}{16.62 \scriptsize{± 4.68}} &
  \multicolumn{3}{c}{9.76 \scriptsize{± 2.76}} \\ \hline
\end{tabular}%
}
\label{tab:multimodal-training}
\end{table*}

\begin{table}[!htb]
\caption{Multimodal results, evaluated on ESC-50 and UrbanSounds8K datasets for the classification task. The evaluated models are categorized as BM (Bimodal Models), MM (Multimodal Models), and CACARA (CACARA Multimodal Models. \textbf{Bolded} results indicate the best performance among multimodal models, while \underline{underlined} results represent the second-best values for the same set.}
\resizebox{0.65\columnwidth}{!}{%
\begin{tabular}{clccllcl}
\hline
\multicolumn{2}{c}{\multirow{3}{*}{Model}}                       & \multicolumn{3}{c}{ESC-50}           & \multicolumn{3}{c}{UrbanSounds8K} \\ \cline{3-8} 
 & & \multicolumn{6}{c}{Accuracy (\%)} \\ \hline
\multirow{3}{*}{\rotatebox[origin=c]{90}{BM}} &
  CLAP (Microsoft) &
  \multicolumn{3}{c}{93.85} &
  \multicolumn{3}{c}{82.74} \\
  & CLAP (LAION)                           & \multicolumn{3}{c}{83.10}            & \multicolumn{3}{c}{80.91}         \\
  & WavCaps                                & \multicolumn{3}{c}{94.25}            & \multicolumn{3}{c}{82.28}         \\ \hline
\multirow{3}{*}{\rotatebox[origin=c]{90}{MM}} &
  ImageBind &
  \multicolumn{3}{c}{64.15} &
  \multicolumn{3}{c}{48.20} \\
  & VAST                                   & \multicolumn{3}{c}{76.80}            & \multicolumn{3}{c}{68.00}         \\
  & LanguageBind                           & \multicolumn{3}{c}{\textbf{94.75}}   & \multicolumn{3}{c}{{\ul 79.24}}   \\ \hline
\multirow{7}{*}{\rotatebox[origin=c]{90}{CACARA}} &
  $\text{CACARA}_{AA/WC/AC/C}$ &
  \multicolumn{3}{c}{89.95 \scriptsize{± 2.14}} &
  \multicolumn{3}{c}{77.04 \scriptsize{± 2.98}} \\
  & $\text{CACARA}_{WC/AC/C}$              & \multicolumn{3}{c}{94.15 \scriptsize{±  0.10}} & \multicolumn{3}{c}{77.40 \scriptsize{± 1.07}}      \\
 &
  $\text{CACARA}_{WC/AC/C/f~0.1}$ &
  \multicolumn{3}{c}{{\ul 94.37} \scriptsize{± 0.49}} &
  \multicolumn{3}{c}{\textbf{79.51} \scriptsize{± 1.06}} \\
  & $\text{CACARA}_{WC/AC/C/f~0.2}$        & \multicolumn{3}{c}{94.00 \scriptsize{± 1.39}} & \multicolumn{3}{c}{77.83 \scriptsize{± 0.11}}      \\
  & $\text{CACARA}_{ASC/AA/WC/AC/C}$       & \multicolumn{3}{c}{82.45 \scriptsize{± 0.70}} & \multicolumn{3}{c}{70.04 \scriptsize{± 1.48}}      \\
  & $\text{CACARA}_{ASC/AA/WC/AC/C/f~0.1}$ & \multicolumn{3}{c}{89.97 \scriptsize{± 0.38}} & \multicolumn{3}{c}{75.40 \scriptsize{± 1.19}}      \\
  & $\text{CACARA}_{ASC/AA/WC/AC/C/f~0.2}$ & \multicolumn{3}{c}{91.35 \scriptsize{± 0.69}} & \multicolumn{3}{c}{74.99 \scriptsize{± 0.55}}      \\ \hline
\end{tabular}%
}
\label{tab:classification-results}
\end{table}

Table~\ref{tab:classification-results} shows the classification results for  ESC-50 and UrbanSounds8K datasets. The performance differences across models in this task are relatively small. For ESC-50, LanguageBind achieved the highest mean accuracy ($94.75\%$), followed closely by CACARA trained with WavCaps ($94.37\%$) and the best-performing bimodal model, WavCaps ($94.25\%$), with a maximum difference of only $0.38$~pp. For UrbanSounds8K, CACARA trained on WavCaps ($79.51\%$) was the best multimodal model, while LanguageBind achieved $79.24\%$, a difference of only $0.27$ pp. However, the bimodal model CLAP (Microsoft) achieved a result of $82.74\%$, demonstrating an advantage in this task and dataset.

\subsection{Multilingual Evaluation}
\label{subsec:multilingual-analysis}

Beyond its multimodal capabilities, CACARA is inherently multilingual, supporting approximately $100$ languages without explicit training for aligning new languages to audio data. We selected twelve languages to evaluate this emergent alignment and analyzed performance in audio-text retrieval and classification tasks. Due to many results in multiple languages, we selected two versions of CACARA for each task to provide a focused comparison.

\begin{table*}[!htb]
\caption{Recall value for the audio-to-text and text-to-audio retrieval tasks on two different CACARA models ($\text{CACARA}_{ASC/AA/WC/AC/C/f~0.2}$ for the AudioCaps dataset and $\text{CACARA}_{WC/AC/C/f~0.2}$ for the ClothoV2 dataset) across the twelve evaluated languages.}
\resizebox{1\textwidth}{!}{%
\begin{tabular}{lrcrrcrrcrrcrrcrrcrrcrrcr}
\hline
& \multicolumn{12}{c|}{Audio-to-Text}                                                                                  & \multicolumn{12}{c}{Text-to-Audio}                                                                                  \\ \hline 
                                 & \multicolumn{24}{c}{$\text{CACARA}_{ASC/AA/WC/AC/C/f~0.2}$ in AudiocCaps}                                                                                                                                                                   \\ \hline 
Language                         & \multicolumn{3}{c}{R@1} & \multicolumn{3}{c}{R@5} & \multicolumn{3}{c}{R@10} & \multicolumn{3}{c|}{R@Avg}            & \multicolumn{3}{c}{R@1} & \multicolumn{3}{c}{R@5} & \multicolumn{3}{c}{R@10} & \multicolumn{3}{c}{R@Avg}            \\ \hline
\multicolumn{1}{l}{English}     & \multicolumn{3}{c}{33.98 \scriptsize{± 0.64}}   & \multicolumn{3}{c}{68.30 \scriptsize{± 0.43}}   & \multicolumn{3}{c}{81.81 \scriptsize{± 0.21}}   & \multicolumn{3}{c|}{61.36 \scriptsize{± 0.27}} & \multicolumn{3}{c}{7.30 \scriptsize{± 0.15}}   & \multicolumn{3}{c}{27.87 \scriptsize{± 0.32}}   & \multicolumn{3}{c}{41.21 \scriptsize{± 0.29}}   & \multicolumn{3}{c}{25.46 \scriptsize{± 0.16}} \\
\multicolumn{1}{l}{Portuguese}  & \multicolumn{3}{c}{21.62 \scriptsize{± 0.70}}   & \multicolumn{3}{c}{51.27 \scriptsize{± 1.31}}   & \multicolumn{3}{c}{66.34 \scriptsize{± 1.23}}   & \multicolumn{3}{c|}{46.41 \scriptsize{± 1.00}} & \multicolumn{3}{c}{5.74 \scriptsize{± 0.18}}   & \multicolumn{3}{c}{21.46 \scriptsize{± 0.18}}   & \multicolumn{3}{c}{33.47 \scriptsize{± 0.35}}   & \multicolumn{3}{c}{20.22 \scriptsize{± 0.11}} \\
\multicolumn{1}{l}{Spanish}     & \multicolumn{3}{c}{24.06 \scriptsize{± 0.48}}   & \multicolumn{3}{c}{53.94 \scriptsize{± 0.87}}   & \multicolumn{3}{c}{68.05 \scriptsize{± 0.35}}   & \multicolumn{3}{c|}{48.68 \scriptsize{± 0.51}} & \multicolumn{3}{c}{5.85 \scriptsize{± 0.16}}   & \multicolumn{3}{c}{22.58 \scriptsize{± 0.29}}   & \multicolumn{3}{c}{34.31 \scriptsize{± 0.25}}   & \multicolumn{3}{c}{20.91 \scriptsize{± 0.08}} \\
\multicolumn{1}{l}{French}      & \multicolumn{3}{c}{21.87 \scriptsize{± 0.70}}   & \multicolumn{3}{c}{51.20 \scriptsize{± 1.51}}   & \multicolumn{3}{c}{66.31 \scriptsize{± 1.29}}   & \multicolumn{3}{c|}{46.46 \scriptsize{± 1.17}} & \multicolumn{3}{c}{5.94 \scriptsize{± 0.05}}   & \multicolumn{3}{c}{22.97 \scriptsize{± 0.27}}   & \multicolumn{3}{c}{35.19 \scriptsize{± 0.47}}   & \multicolumn{3}{c}{21.37 \scriptsize{± 0.16}} \\
\multicolumn{1}{l}{Russian}     & \multicolumn{3}{c}{20.73 \scriptsize{± 0.47}}   & \multicolumn{3}{c}{48.96 \scriptsize{± 0.60}}   & \multicolumn{3}{c}{63.03 \scriptsize{± 0.74}}   & \multicolumn{3}{c|}{44.24 \scriptsize{± 0.58}} & \multicolumn{3}{c}{5.55 \scriptsize{± 0.31}}   & \multicolumn{3}{c}{19.48 \scriptsize{± 0.47}}   & \multicolumn{3}{c}{30.70 \scriptsize{± 0.39}}   & \multicolumn{3}{c}{18.58 \scriptsize{± 0.39}} \\
\multicolumn{1}{l}{Arabic}      & \multicolumn{3}{c}{15.25 \scriptsize{± 0.73}}   & \multicolumn{3}{c}{39.70 \scriptsize{± 1.05}}   & \multicolumn{3}{c}{53.90 \scriptsize{± 1.15}}   & \multicolumn{3}{c|}{36.28 \scriptsize{± 0.97}} & \multicolumn{3}{c}{4.71 \scriptsize{± 0.15}}   & \multicolumn{3}{c}{18.45 \scriptsize{± 0.23}}   & \multicolumn{3}{c}{28.27 \scriptsize{± 0.31}}   & \multicolumn{3}{c}{17.15 \scriptsize{± 0.10}} \\
\multicolumn{1}{l}{Hindi}       & \multicolumn{3}{c}{14.08 \scriptsize{± 0.41}}   & \multicolumn{3}{c}{37.47 \scriptsize{± 0.30}}   & \multicolumn{3}{c}{51.51 \scriptsize{± 0.80}}   & \multicolumn{3}{c|}{34.35 \scriptsize{± 0.42}} & \multicolumn{3}{c}{3.71 \scriptsize{± 0.08}}   & \multicolumn{3}{c}{14.97 \scriptsize{± 0.09}}   & \multicolumn{3}{c}{23.56 \scriptsize{± 0.07}}   & \multicolumn{3}{c}{14.08 \scriptsize{± 0.02}} \\
\multicolumn{1}{l}{German}      & \multicolumn{3}{c}{23.95 \scriptsize{± 0.41}}   & \multicolumn{3}{c}{54.36 \scriptsize{± 0.06}}   & \multicolumn{3}{c}{68.26 \scriptsize{± 0.14}}   & \multicolumn{3}{c|}{48.86 \scriptsize{± 0.12}} & \multicolumn{3}{c}{6.00 \scriptsize{± 0.09}}   & \multicolumn{3}{c}{23.09 \scriptsize{± 0.37}}   & \multicolumn{3}{c}{35.04 \scriptsize{± 0.45}}   & \multicolumn{3}{c}{21.38 \scriptsize{± 0.25}} \\
\multicolumn{1}{l}{Chinese(zh)} & \multicolumn{3}{c}{18.24 \scriptsize{± 0.28}}   & \multicolumn{3}{c}{46.59 \scriptsize{± 1.11}}   & \multicolumn{3}{c}{62.04 \scriptsize{± 0.42}}   & \multicolumn{3}{c|}{42.29 \scriptsize{± 0.47}} & \multicolumn{3}{c}{5.45 \scriptsize{± 0.12}}   & \multicolumn{3}{c}{20.52 \scriptsize{± 0.20}}   & \multicolumn{3}{c}{31.70 \scriptsize{± 0.23}}   & \multicolumn{3}{c}{19.23 \scriptsize{± 0.09}} \\
\multicolumn{1}{l}{Swahili}     & \multicolumn{3}{c}{1.11  \scriptsize{± 0.23}}   & \multicolumn{3}{c}{3.96  \scriptsize{± 0.11}}   & \multicolumn{3}{c}{6.48  \scriptsize{± 0.24}}   & \multicolumn{3}{c|}{3.85  \scriptsize{± 0.19}} & \multicolumn{3}{c}{0.65 \scriptsize{± 0.07}}   & \multicolumn{3}{c}{2.29  \scriptsize{± 0.06}}   & \multicolumn{3}{c}{3.67  \scriptsize{± 0.08}}   & \multicolumn{3}{c}{2.20  \scriptsize{± 0.02}} \\
\multicolumn{1}{l}{Japanese}    & \multicolumn{3}{c}{21.36 \scriptsize{± 0.49}}   & \multicolumn{3}{c}{51.46 \scriptsize{± 0.24}}   & \multicolumn{3}{c}{65.64 \scriptsize{± 0.38}}   & \multicolumn{3}{c|}{46.15 \scriptsize{± 0.22}} & \multicolumn{3}{c}{5.59 \scriptsize{± 0.24}}   & \multicolumn{3}{c}{21.87 \scriptsize{± 0.48}}   & \multicolumn{3}{c}{33.82 \scriptsize{± 0.20}}   & \multicolumn{3}{c}{20.42 \scriptsize{± 0.30}} \\
\multicolumn{1}{l}{Turkish}     & \multicolumn{3}{c}{17.04 \scriptsize{± 0.40}}   & \multicolumn{3}{c}{42.65 \scriptsize{± 0.63}}   & \multicolumn{3}{c}{56.80 \scriptsize{± 0.45}}  & \multicolumn{3}{c|}{38.84 \scriptsize{± 0.47}} & \multicolumn{3}{c}{4.42 \scriptsize{± 0.23}}   & \multicolumn{3}{c}{17.22 \scriptsize{± 0.32}}   & \multicolumn{3}{c}{27.15 \scriptsize{± 0.51}}   & \multicolumn{3}{c}{16.27 \scriptsize{± 0.33}} \\ \hline
                                 & \multicolumn{24}{c}{$\text{CACARA}_{WC/AC/C/f~0.2}$ in ClothoV2}                                                                                                                                                                            \\ \hline 
                                 & \multicolumn{3}{c}{R@1} & \multicolumn{3}{c}{R@5} & \multicolumn{3}{c}{R@10} & \multicolumn{3}{r|}{R@Avg}            & \multicolumn{3}{c}{R@1} & \multicolumn{3}{c}{R@5} & \multicolumn{3}{c}{R@10} & \multicolumn{3}{c}{R@Avg}             \\ \hline
English                          & \multicolumn{3}{c}{17.26 \scriptsize{± 2.33}}   & \multicolumn{3}{c}{40.91 \scriptsize{± 5.84}}   & \multicolumn{3}{c}{53.85 \scriptsize{± 6.74}}   & \multicolumn{3}{c|}{37.34 \scriptsize{± 4.96}} & \multicolumn{3}{c}{4.03 \scriptsize{± 0.36}}   & \multicolumn{3}{c}{15.60 \scriptsize{± 1.25}}   & \multicolumn{3}{c}{24.78 \scriptsize{± 2.36}}   & \multicolumn{3}{c}{14.80 \scriptsize{± 1.32}}                      \\
Portuguese                       & \multicolumn{3}{c}{10.83 \scriptsize{± 0.60}}   & \multicolumn{3}{c}{28.98 \scriptsize{± 1.56}}   & \multicolumn{3}{c}{39.72 \scriptsize{± 1.93}}   & \multicolumn{3}{c|}{26.51 \scriptsize{± 1.35}} & \multicolumn{3}{c}{3.09 \scriptsize{± 0.16}}   & \multicolumn{3}{c}{11.61 \scriptsize{± 0.09}}   & \multicolumn{3}{c}{18.57 \scriptsize{± 0.12}}   & \multicolumn{3}{c}{11.09 \scriptsize{± 0.05}}                      \\
Spanish                          & \multicolumn{3}{c}{11.37 \scriptsize{± 0.39}}   & \multicolumn{3}{c}{29.88 \scriptsize{± 1.73}}   & \multicolumn{3}{c}{40.54 \scriptsize{± 2.09}}   & \multicolumn{3}{c|}{27.26 \scriptsize{± 1.34}} & \multicolumn{3}{c}{3.18 \scriptsize{± 0.12}}   & \multicolumn{3}{c}{12.09 \scriptsize{± 0.17}}   & \multicolumn{3}{c}{19.38 \scriptsize{± 0.05}}   & \multicolumn{3}{c}{11.55 \scriptsize{± 0.05}}                      \\
French                           & \multicolumn{3}{c}{10.83 \scriptsize{± 0.66}}   & \multicolumn{3}{c}{28.75 \scriptsize{± 1.43}}   & \multicolumn{3}{c}{39.55 \scriptsize{± 1.35}}   & \multicolumn{3}{c|}{26.38 \scriptsize{± 1.10}} & \multicolumn{3}{c}{3.00 \scriptsize{± 0.14}}   & \multicolumn{3}{c}{11.50 \scriptsize{± 0.23}}   & \multicolumn{3}{c}{18.05 \scriptsize{± 0.12}}   & \multicolumn{3}{c}{10.85 \scriptsize{± 0.12}}                      \\
Russian                          & \multicolumn{3}{c}{8.94  \scriptsize{± 0.51}}   & \multicolumn{3}{c}{24.45 \scriptsize{± 1.88}}   & \multicolumn{3}{c}{34.62 \scriptsize{± 2.55}}   & \multicolumn{3}{c|}{22.67 \scriptsize{± 1.64}} & \multicolumn{3}{c}{2.78 \scriptsize{± 0.14}}   & \multicolumn{3}{c}{10.60 \scriptsize{± 0.33}}   & \multicolumn{3}{c}{16.82 \scriptsize{± 0.51}}   & \multicolumn{3}{c}{10.06 \scriptsize{± 0.24}}                      \\
Arabic                           & \multicolumn{3}{c}{6.84  \scriptsize{± 0.61}}   & \multicolumn{3}{c}{20.34 \scriptsize{± 1.22}}   & \multicolumn{3}{c}{29.13 \scriptsize{± 1.31}}   & \multicolumn{3}{c|}{18.77 \scriptsize{± 1.02}} & \multicolumn{3}{c}{2.34 \scriptsize{± 0.08}}   & \multicolumn{3}{c}{8.85  \scriptsize{± 0.04}}   & \multicolumn{3}{c}{14.50 \scriptsize{± 0.08}}   & \multicolumn{3}{c}{8.56  \scriptsize{± 0.05}}                      \\
Hindi                            & \multicolumn{3}{c}{6.02  \scriptsize{± 0.43}}   & \multicolumn{3}{c}{17.30 \scriptsize{± 1.24}}   & \multicolumn{3}{c}{25.69 \scriptsize{± 1.60}}   & \multicolumn{3}{c|}{16.34 \scriptsize{± 1.05}} & \multicolumn{3}{c}{2.04 \scriptsize{± 0.16}}   & \multicolumn{3}{c}{7.80  \scriptsize{± 0.30}}   & \multicolumn{3}{c}{12.33 \scriptsize{± 0.54}}   & \multicolumn{3}{c}{7.39  \scriptsize{± 0.31}}                      \\
German                           & \multicolumn{3}{c}{11.55 \scriptsize{± 0.72}}   & \multicolumn{3}{c}{29.99 \scriptsize{± 1.50}}   & \multicolumn{3}{c}{40.96 \scriptsize{± 1.49}}   & \multicolumn{3}{c|}{27.49 \scriptsize{± 1.21}} & \multicolumn{3}{c}{3.21 \scriptsize{± 0.15}}   & \multicolumn{3}{c}{12.45 \scriptsize{± 0.01}}   & \multicolumn{3}{c}{19.55 \scriptsize{± 0.27}}   & \multicolumn{3}{c}{11.74 \scriptsize{± 0.13}}                      \\
Chinese(zh)                      & \multicolumn{3}{c}{9.55  \scriptsize{± 0.38}}   & \multicolumn{3}{c}{25.94 \scriptsize{± 1.73}}   & \multicolumn{3}{c}{36.59 \scriptsize{± 1.80}}   & \multicolumn{3}{c|}{24.03 \scriptsize{± 1.29}} & \multicolumn{3}{c}{2.79 \scriptsize{± 0.17}}   & \multicolumn{3}{c}{11.02 \scriptsize{± 0.15}}   & \multicolumn{3}{c}{17.64 \scriptsize{± 0.43}}   & \multicolumn{3}{c}{10.48 \scriptsize{± 0.19}}                      \\
Swahili                          & \multicolumn{3}{c}{1.03  \scriptsize{± 0.03}}   & \multicolumn{3}{c}{3.35 \scriptsize{± 0.16}}   & \multicolumn{3}{c}{5.39  \scriptsize{± 0.27}}   & \multicolumn{3}{c|}{3.26  \scriptsize{± 0.14}} & \multicolumn{3}{c}{0.52 \scriptsize{± 0.05}}   & \multicolumn{3}{c}{1.70  \scriptsize{± 0.13}}   & \multicolumn{3}{c}{2.65  \scriptsize{± 0.20}}   & \multicolumn{3}{c}{1.62  \scriptsize{± 0.09}}                      \\
Japanese                         & \multicolumn{3}{c}{10.75 \scriptsize{± 0.73}}   & \multicolumn{3}{c}{29.19 \scriptsize{± 1.89}}   & \multicolumn{3}{c}{40.24 \scriptsize{± 2.10}}   & \multicolumn{3}{c|}{26.72 \scriptsize{± 1.55}} & \multicolumn{3}{c}{2.97 \scriptsize{± 0.19}}   & \multicolumn{3}{c}{11.50 \scriptsize{± 0.37}}   & \multicolumn{3}{c}{18.32 \scriptsize{± 0.34}}   & \multicolumn{3}{c}{10.93 \scriptsize{± 0.22}}                      \\
Turkish                          & \multicolumn{3}{c}{7.93  \scriptsize{± 0.74}}   & \multicolumn{3}{c}{22.53 \scriptsize{± 1.23}}   & \multicolumn{3}{c}{32.12 \scriptsize{± 2.69}}   & \multicolumn{3}{c|}{20.86 \scriptsize{± 1.54}} & \multicolumn{3}{c}{2.51 \scriptsize{± 0.07}}   & \multicolumn{3}{c}{9.61  \scriptsize{± 0.20}}   & \multicolumn{3}{c}{15.51 \scriptsize{± 0.37}}   & \multicolumn{3}{c}{9.21  \scriptsize{± 0.17}}                      \\ \hline
\end{tabular}%
}
\label{tab:retrieval-multilingual}
\end{table*}

For the retrieval task, Table~\ref{tab:retrieval-multilingual} presents results for the models $\text{CACARA}_{ASC/AA/WC/AC/C/f~0.2}$ in the AudioCaps dataset and $\text{CACARA}_{W C/AC/C/f~0.2}$ in the ClothoV2 dataset. The audio encoder was trained only in English, resulting in better performance with that language. Therefore, we use English results as the upper bound. The results vary depending on the target language, as factors such as translation quality, linguistic complexity, and modality-specific nuances can significantly influence retrieval performance. 

Some languages---Spanish, German, Portuguese, French, Russian, and Japanese---perform well, achieving R@1 above $20$ for the $\text{CACARA}_{ASC/AA/WC/AC/C/f~0.2}$ model on AudioCaps. On average, the other languages achieve R@1 above $13$, with performance variations primarily influenced by the quantity and quality of textual data used during the pretraining of the text model~\cite{geigle2024babelimagenetmassivelymultilingualevaluation}. This suggests that improving low-resource language performance does not require retraining the entire model. Only Swahili was below average because it has very few resources available. We observed the same behavior from the $\text{CACARA}_{WC/AC/C/f~0.2}$ model in ClothoV2.

For the classification task, Table~\ref{tab:classification_multilingual}~shows results for two models, $\text{CACARA}_{WC/AC/C/f~0.1}$ and $\text{CACARA}_{ASC/AA/WC/AC/C/f~0.2}$, evaluated on ESC-50 and UrbanSounds8K. As expected, languages with more resources showed strong classification performance, similar to retrieval results. In addition, Mandarin showed better results than in the previous task, while Portuguese showed a drop. However, the overall classification average of the different languages is $66.5\%$ for $\text{CACARA}_{WC/AC/C}$ using ESC-50.

\begin{table}[!htb]
\caption{Classification retults on two different CACARA models ($\text{CACARA}_{WC/AC/C/f~0.1}$ and $\text{CACARA}_{ASC/AA/WC/AC/C/f~0.2}$) in the datasets ESC-50 and UrbanSounds8K across the 12~evaluated languages.}
\resizebox{0.9\columnwidth}{!}{%
\begin{tabular}{lccrrcr|ccrccr}
\hline
            & \multicolumn{6}{c|}{$\text{CACARA}_{WC/AC/C/f~0.1}$}     & \multicolumn{6}{c}{$\text{CACARA}_{ASC/AA/WC/AC/C/f~0.2}$} \\ \hline
\multicolumn{1}{l}{Language} & \multicolumn{3}{c|}{ESC-50} & \multicolumn{3}{l|}{UrbanSounds8K} & \multicolumn{3}{c|}{ESC-50} & \multicolumn{3}{c}{UrbanSounds8K} \\ \hline
English     & \multicolumn{3}{c|}{94.37 \scriptsize{± 0.49}} & \multicolumn{3}{c|}{79.51 \scriptsize{± 1.06}} & \multicolumn{3}{c|}{91.35 \scriptsize{± 0.69}} & \multicolumn{3}{c}{74.99 \scriptsize{± 0.55}} \\
Portuguese  & \multicolumn{3}{c|}{79.63 \scriptsize{± 0.81}} & \multicolumn{3}{c|}{66.49 \scriptsize{± 0.79}} & \multicolumn{3}{c|}{80.92 \scriptsize{± 0.98}} & \multicolumn{3}{c}{71.38 \scriptsize{± 1.35}} \\
Spanish     & \multicolumn{3}{c|}{86.25 \scriptsize{± 0.40}} & \multicolumn{3}{c|}{72.02 \scriptsize{± 1.78}} & \multicolumn{3}{c|}{85.60 \scriptsize{± 0.78}} & \multicolumn{3}{c}{71.02 \scriptsize{± 0.67}} \\
French      & \multicolumn{3}{c|}{83.60 \scriptsize{± 0.69}} & \multicolumn{3}{c|}{69.56 \scriptsize{± 0.65}} & \multicolumn{3}{c|}{82.45 \scriptsize{± 0.49}} & \multicolumn{3}{c}{67.72 \scriptsize{± 1.12}} \\
Russian     & \multicolumn{3}{c|}{81.15 \scriptsize{± 0.33}} & \multicolumn{3}{c|}{71.77 \scriptsize{± 1.00}} & \multicolumn{3}{c|}{78.30 \scriptsize{± 1.08}} & \multicolumn{3}{c}{67.16 \scriptsize{± 1.32}} \\
Arabic      & \multicolumn{3}{c|}{65.28 \scriptsize{± 0.95}} & \multicolumn{3}{c|}{63.13 \scriptsize{± 0.64}} & \multicolumn{3}{c|}{63.65 \scriptsize{± 1.09}} & \multicolumn{3}{c}{63.77 \scriptsize{± 0.60}} \\
Hindi       & \multicolumn{3}{c|}{64.30 \scriptsize{± 0.97}} & \multicolumn{3}{c|}{58.19 \scriptsize{± 0.84}} & \multicolumn{3}{c|}{60.12 \scriptsize{± 0.83}} & \multicolumn{3}{c}{58.11 \scriptsize{± 1.15}} \\
German      & \multicolumn{3}{c|}{82.12 \scriptsize{± 0.87}} & \multicolumn{3}{c|}{74.99 \scriptsize{± 0.12}} & \multicolumn{3}{c|}{75.92 \scriptsize{± 1.08}} & \multicolumn{3}{c}{72.38 \scriptsize{± 0.10}} \\
Chinese(zh) & \multicolumn{3}{c|}{83.47 \scriptsize{± 1.68}} & \multicolumn{3}{c|}{70.14 \scriptsize{± 0.12}} & \multicolumn{3}{c|}{81.23 \scriptsize{± 1.07}} & \multicolumn{3}{c}{67.22 \scriptsize{± 1.15}} \\
Swahili     & \multicolumn{3}{c|}{20.55 \scriptsize{± 0.41}} & \multicolumn{3}{c|}{42.99 \scriptsize{± 1.08}} & \multicolumn{3}{c|}{20.48 \scriptsize{± 2.17}} & \multicolumn{3}{c}{38.35 \scriptsize{± 1.97}} \\
Japanese    & \multicolumn{3}{c|}{84.08 \scriptsize{± 1.63}} & \multicolumn{3}{c|}{68.79 \scriptsize{± 0.02}} & \multicolumn{3}{c|}{81.78 \scriptsize{± 0.75}} & \multicolumn{3}{c}{73.03 \scriptsize{± 0.74}} \\
Turkish     & \multicolumn{3}{c|}{74.12 \scriptsize{± 1.83}} & \multicolumn{3}{c|}{64.94 \scriptsize{± 0.68}} & \multicolumn{3}{c|}{69.85 \scriptsize{± 1.97}} & \multicolumn{3}{c}{63.91 \scriptsize{± 0.82}} \\ \hline
\end{tabular}%
}
\label{tab:classification_multilingual}
\end{table}

\subsection{Efficiency and Scaling}
\label{subsec:compute-resources}

To quantify the computational savings achieved by our approach, we conducted a comparative analysis between the proposed CACARA method and a fully trained tri-modal baseline using a 5k-sample subset from the VAST~\cite{chen2023vast} dataset. The baseline trained text, vision, and audio encoders concurrently, whereas CACARA froze the text and vision encoders and tuned only the audio encoder. Both methods were trained for $100$ epochs across three independent runs to account for variance.

We measured key metrics, including floating-point operations (GFLOPs), multiply-accumulate operations (GMACs), parameter count, total training time, energy consumption~(kWh), and estimated carbon emissions (CO$_2$e), as presented in Table~\ref{tab:computer-savings}.

\begin{table}[!htb]
\caption{Computational efficiency and environmental impact comparison between the proposed CACARA model, which tunes only the audio encoder, and a fully tri-modal baseline. Results are averaged over three runs on a 5k-sample VAST dataset subset.}
\resizebox{\columnwidth}{!}{%
\begin{tabular}{lcccccc}
\hline
\multicolumn{1}{c}{Model} & GFLOPs & GMACS  & Parameters & Time (hh:mm)                & Energy (kWh) & Emission (CO$_2$e)    \\ \hline
CACARA                    & 185,00 & 92,42  & 369,40     & 04:47 \scriptsize{± 00:05} & 1,91 \scriptsize{± 0,03}  & 0,15 \scriptsize{± 0}    \\
Fully Tri-Modal           & 246,12 & 122,95 & 457,25     & 25:12 \scriptsize{± 2:11}  & 6,66 \scriptsize{± 0,43}  & 0,51 \scriptsize{± 0,03} \\ \hline
Reduction (\%)                 & 25    & 25    & 19        & 79                 & 73          & 73         \\ \hline
\end{tabular}%
}
\label{tab:computer-savings}
\end{table}



As shown, CACARA reduces computational requirements (GFLOPs and GMACs) by $25$~pp and the parameter count by $19$~pp. This results in a substantial $79$~pp reduction in total training time, accompanied by a $73$~pp decrease in both energy consumption and carbon emissions. These findings underscore the significant efficiency gains and environmental benefits afforded by our~approach.

\subsection{Expanded Resources}
\label{subsec:extended-resources}

To achieve a single robust model applicable across tasks, we trained an expanded-resource version of CACARA: $\text{CACARA}_{ASC/AA/WC/AC/C/f~0.2}$. Unlike previous models trained with $3$~epochs, a batch size of $64$, and different datasets, we trained this version with $10$ epochs, a batch size of $110$,  and all datasets together, with a data filtering of $0.2$, for more complete training.

A direct comparison of this expanded model against the best-performing optimized models is in Table~\ref{tab:retrieval_expanded_resources}, evaluating audio-to-text and text-to-audio retrieval across $12$ languages. This task in the AudioCaps dataset did not improve when given more resources and training time and continues to show values lower than those obtained with the optimized model. This is due to the proximity of the trained sets to the distribution of the test set. However, it still improved over the same model with fewer resources. For the same task, in the ClothoV2 dataset, a general improvement is observed for the audio-to-text retrieval task, but text-to-audio retrieval remained below the optimized model’s results.

\begin{table*}[!htb]
\caption{Retrieval task results for expanded resources experiments, divided into two datasets: AudioCaps and ClothoV2, and two retrieval tasks: audio-to-text and text-to-audio across the 12~evaluated languages.}
\resizebox{\textwidth}{!}{%
\begin{tabular}{lcccccccccccccccccccccccccccccccccccccccccccccccc}
\hline
 &
  \multicolumn{24}{c|}{Optimized Basic Model} &
  \multicolumn{24}{c}{Expanded Resources Model} \\ \hline
 &
  \multicolumn{12}{|c|}{Audio-to-Text} &
  \multicolumn{12}{c|}{Text-to-Audio} &
  \multicolumn{12}{c|}{Audio-to-Text} &
  \multicolumn{12}{c}{Text-to-Audio} \\ \hline
 &
  \multicolumn{48}{c}{AudioCaps} \\ \hline
Language &
  \multicolumn{3}{|c}{R@1} &
  \multicolumn{3}{c}{R@5} &
  \multicolumn{3}{c}{R@10} &
  \multicolumn{3}{c|}{R@Avg} &
  \multicolumn{3}{c}{R@1} &
  \multicolumn{3}{c}{R@5} &
  \multicolumn{3}{c}{R@10} &
  \multicolumn{3}{c|}{R@Avg} &
  \multicolumn{3}{c}{R@1} &
  \multicolumn{3}{c}{R@5} &
  \multicolumn{3}{c}{R@10} &
  \multicolumn{3}{c|}{R@Avg} &
  \multicolumn{3}{c}{R@1} &
  \multicolumn{3}{c}{R@5} &
  \multicolumn{3}{c}{R@10} &
  \multicolumn{3}{c}{R@Avg} \\ \hline
\multicolumn{1}{l|}{English} &
  \multicolumn{3}{c}{33.98 \scriptsize{± 0.64}} &
  \multicolumn{3}{c}{68.30 \scriptsize{± 0.43}} &
  \multicolumn{3}{c}{81.81 \scriptsize{± 0.21}} &
  \multicolumn{3}{c|}{61.36 \scriptsize{± 0.27}} &
  \multicolumn{3}{c}{7.30 \scriptsize{± 0.15}} &
  \multicolumn{3}{c}{27.87 \scriptsize{± 0.32}} &
  \multicolumn{3}{c}{41.21 \scriptsize{± 0.29}} &
  \multicolumn{3}{c|}{25.46 \scriptsize{± 0.16}} &
  \multicolumn{3}{c}{31.45 \scriptsize{± 0.14}} &
  \multicolumn{3}{c}{66.09 \scriptsize{± 0.31}} &
  \multicolumn{3}{c}{79.61 \scriptsize{± 0.45}} &
  \multicolumn{3}{c|}{59.05 \scriptsize{± 0.28}} &
  \multicolumn{3}{c}{7.64 \scriptsize{± 0.09}} &
  \multicolumn{3}{c}{28.78 \scriptsize{± 0.21}} &
  \multicolumn{3}{c}{42.95 \scriptsize{± 0.14}} &
  \multicolumn{3}{c}{26.46 \scriptsize{± 0.12}} \\
\multicolumn{1}{l|}{Portuguese} &
  \multicolumn{3}{c}{21.62 \scriptsize{± 0.70}} &
  \multicolumn{3}{c}{51.27 \scriptsize{± 1.31}} &
  \multicolumn{3}{c}{66.34 \scriptsize{± 1.23}} &
  \multicolumn{3}{c|}{46.41 \scriptsize{± 1.00}} &
  \multicolumn{3}{c}{5.74 \scriptsize{± 0.18}} &
  \multicolumn{3}{c}{21.46 \scriptsize{± 0.18}} &
  \multicolumn{3}{c}{33.47 \scriptsize{± 0.35}} &
  \multicolumn{3}{c|}{20.22 \scriptsize{± 0.11}} &
  \multicolumn{3}{c}{18.54 \scriptsize{± 0.98}} &
  \multicolumn{3}{c}{46.57 \scriptsize{± 0.75}} &
  \multicolumn{3}{c}{60.96 \scriptsize{± 0.53}} &
  \multicolumn{3}{c|}{42.02 \scriptsize{± 0.74}} &
  \multicolumn{3}{c}{5.99 \scriptsize{± 0.10}} &
  \multicolumn{3}{c}{22.07 \scriptsize{± 0.43}} &
  \multicolumn{3}{c}{33.75 \scriptsize{± 0.06}} &
  \multicolumn{3}{c}{20.60 \scriptsize{± 0.19}} \\
\multicolumn{1}{l|}{Spanish} &
  \multicolumn{3}{c}{24.06 \scriptsize{± 0.48}} &
  \multicolumn{3}{c}{53.94 \scriptsize{± 0.87}} &
  \multicolumn{3}{c}{68.05 \scriptsize{± 0.35}} &
  \multicolumn{3}{c|}{48.68 \scriptsize{± 0.51}} &
  \multicolumn{3}{c}{5.85 \scriptsize{± 0.16}} &
  \multicolumn{3}{c}{22.58 \scriptsize{± 0.29}} &
  \multicolumn{3}{c}{34.31 \scriptsize{± 0.25}} &
  \multicolumn{3}{c|}{20.91 \scriptsize{± 0.08}} &
  \multicolumn{3}{c}{21.03 \scriptsize{± 0.38}} &
  \multicolumn{3}{c}{50.85 \scriptsize{± 0.26}} &
  \multicolumn{3}{c}{65.32 \scriptsize{± 0.80}} &
  \multicolumn{3}{c|}{45.73 \scriptsize{± 0.45}} &
  \multicolumn{3}{c}{6.04 \scriptsize{± 0.10}} &
  \multicolumn{3}{c}{22.46 \scriptsize{± 0.21}} &
  \multicolumn{3}{c}{34.62 \scriptsize{± 0.05}} &
  \multicolumn{3}{c}{21.04 \scriptsize{± 0.06}} \\
\multicolumn{1}{l|}{French} &
  \multicolumn{3}{c}{21.87 \scriptsize{± 0.70}} &
  \multicolumn{3}{c}{51.20 \scriptsize{± 1.51}} &
  \multicolumn{3}{c}{66.31 \scriptsize{± 1.29}} &
  \multicolumn{3}{c|}{46.46 \scriptsize{± 1.17}} &
  \multicolumn{3}{c}{5.94 \scriptsize{± 0.05}} &
  \multicolumn{3}{c}{22.97 \scriptsize{± 0.27}} &
  \multicolumn{3}{c}{35.19 \scriptsize{± 0.47}} &
  \multicolumn{3}{c|}{21.37 \scriptsize{± 0.16}} &
  \multicolumn{3}{c}{20.33 \scriptsize{± 0.42}} &
  \multicolumn{3}{c}{48.61 \scriptsize{± 0.36}} &
  \multicolumn{3}{c}{62.67 \scriptsize{± 0.38}} &
  \multicolumn{3}{c|}{43.87 \scriptsize{± 0.27}} &
  \multicolumn{3}{c}{6.25 \scriptsize{± 0.08}} &
  \multicolumn{3}{c}{23.44 \scriptsize{± 0.22}} &
  \multicolumn{3}{c}{35.66 \scriptsize{± 0.23}} &
  \multicolumn{3}{c}{21.78 \scriptsize{± 0.06}} \\
\multicolumn{1}{l|}{Russian} &
  \multicolumn{3}{c}{20.73 \scriptsize{± 0.47}} &
  \multicolumn{3}{c}{48.96 \scriptsize{± 0.60}} &
  \multicolumn{3}{c}{63.03 \scriptsize{± 0.74}} &
  \multicolumn{3}{c|}{44.24 \scriptsize{± 0.58}} &
  \multicolumn{3}{c}{5.55 \scriptsize{± 0.31}} &
  \multicolumn{3}{c}{19.48 \scriptsize{± 0.47}} &
  \multicolumn{3}{c}{30.70 \scriptsize{± 0.39}} &
  \multicolumn{3}{c|}{18.58 \scriptsize{± 0.39}} &
  \multicolumn{3}{c}{17.58 \scriptsize{± 0.35}} &
  \multicolumn{3}{c}{44.19 \scriptsize{± 1.45}} &
  \multicolumn{3}{c}{58.16 \scriptsize{± 0.89}} &
  \multicolumn{3}{c|}{39.97 \scriptsize{± 0.89}} &
  \multicolumn{3}{c}{5.50 \scriptsize{± 0.32}} &
  \multicolumn{3}{c}{19.88 \scriptsize{± 0.34}} &
  \multicolumn{3}{c}{31.28 \scriptsize{± 0.42}} &
  \multicolumn{3}{c}{18.89 \scriptsize{± 0.30}} \\
\multicolumn{1}{l|}{Arabic} &
  \multicolumn{3}{c}{15.25 \scriptsize{± 0.73}} &
  \multicolumn{3}{c}{39.70 \scriptsize{± 1.05}} &
  \multicolumn{3}{c}{53.90 \scriptsize{± 1.15}} &
  \multicolumn{3}{c|}{6.28 \scriptsize{± 0.97}} &
  \multicolumn{3}{c}{4.71 \scriptsize{± 0.15}} &
  \multicolumn{3}{c}{18.45 \scriptsize{± 0.23}} &
  \multicolumn{3}{c}{28.27 \scriptsize{± 0.31}} &
  \multicolumn{3}{c|}{17.15 \scriptsize{± 0.10}} &
  \multicolumn{3}{c}{11.18 \scriptsize{± 0.47}} &
  \multicolumn{3}{c}{31.96 \scriptsize{± 0.86}} &
  \multicolumn{3}{c}{45.17 \scriptsize{± 0.51}} &
  \multicolumn{3}{c|}{29.44 \scriptsize{± 0.58}} &
  \multicolumn{3}{c}{4.71 \scriptsize{± 0.17}} &
  \multicolumn{3}{c}{18.37 \scriptsize{± 0.19}} &
  \multicolumn{3}{c}{28.10 \scriptsize{± 0.18}} &
  \multicolumn{3}{c}{17.06 \scriptsize{± 0.07}} \\
\multicolumn{1}{l|}{Hindi} &
  \multicolumn{3}{c}{14.08 \scriptsize{± 0.41}} &
  \multicolumn{3}{c}{37.47 \scriptsize{± 0.30}} &
  \multicolumn{3}{c}{51.51 \scriptsize{± 0.80}} &
  \multicolumn{3}{c|}{34.35 \scriptsize{± 0.42}} &
  \multicolumn{3}{c}{3.71 \scriptsize{± 0.08}} &
  \multicolumn{3}{c}{14.97 \scriptsize{± 0.09}} &
  \multicolumn{3}{c}{23.56 \scriptsize{± 0.07}} &
  \multicolumn{3}{c|}{14.08 \scriptsize{± 0.02}} &
  \multicolumn{3}{c}{11.70 \scriptsize{± 0.30}} &
  \multicolumn{3}{c}{33.33 \scriptsize{± 0.71}} &
  \multicolumn{3}{c}{46.76 \scriptsize{± 0.76}} &
  \multicolumn{3}{c|}{30.60 \scriptsize{± 0.38}} &
  \multicolumn{3}{c}{3.83 \scriptsize{± 0.11}} &
  \multicolumn{3}{c}{15.10 \scriptsize{± 0.44}} &
  \multicolumn{3}{c}{23.28 \scriptsize{± 0.23}} &
  \multicolumn{3}{c}{14.07 \scriptsize{± 0.24}} \\
\multicolumn{1}{l|}{German} &
  \multicolumn{3}{c}{23.95 \scriptsize{± 0.41}} &
  \multicolumn{3}{c}{54.36 \scriptsize{± 0.06}} &
  \multicolumn{3}{c}{68.26 \scriptsize{± 0.14}} &
  \multicolumn{3}{c|}{48.86 \scriptsize{± 0.12}} &
  \multicolumn{3}{c}{6.00 \scriptsize{± 0.09}} &
  \multicolumn{3}{c}{23.09 \scriptsize{± 0.37}} &
  \multicolumn{3}{c}{35.04 \scriptsize{± 0.45}} &
  \multicolumn{3}{c|}{21.38 \scriptsize{± 0.25}} &
  \multicolumn{3}{c}{21.33 \scriptsize{± 0.13}} &
  \multicolumn{3}{c}{50.33 \scriptsize{± 0.90}} &
  \multicolumn{3}{c}{64.41 \scriptsize{± 0.90}} &
  \multicolumn{3}{c|}{45.36 \scriptsize{± 0.62}} &
  \multicolumn{3}{c}{5.85 \scriptsize{± 0.08}} &
  \multicolumn{3}{c}{23.06 \scriptsize{± 0.38}} &
  \multicolumn{3}{c}{35.64 \scriptsize{± 0.46}} &
  \multicolumn{3}{c}{21.52 \scriptsize{± 0.30}} \\
\multicolumn{1}{l|}{Chinese(zh)} &
  \multicolumn{3}{c}{18.24 \scriptsize{± 0.28}} &
  \multicolumn{3}{c}{46.59 \scriptsize{± 1.11}} &
  \multicolumn{3}{c}{62.04 \scriptsize{± 0.42}} &
  \multicolumn{3}{c|}{42.29 \scriptsize{± 0.47}} &
  \multicolumn{3}{c}{5.45 \scriptsize{± 0.12}} &
  \multicolumn{3}{c}{20.52 \scriptsize{± 0.20}} &
  \multicolumn{3}{c}{31.70 \scriptsize{± 0.23}} &
  \multicolumn{3}{c|}{19.23 \scriptsize{± 0.09}} &
  \multicolumn{3}{c}{15.27 \scriptsize{± 0.74}} &
  \multicolumn{3}{c}{41.17 \scriptsize{± 1.37}} &
  \multicolumn{3}{c}{56.36 \scriptsize{± 0.43}} &
  \multicolumn{3}{c|}{37.60 \scriptsize{± 0.80}} &
  \multicolumn{3}{c}{5.57 \scriptsize{± 0.31}} &
  \multicolumn{3}{c}{20.79 \scriptsize{± 0.16}} &
  \multicolumn{3}{c}{32.16 \scriptsize{± 0.52}} &
  \multicolumn{3}{c}{19.51 \scriptsize{± 0.05}} \\
\multicolumn{1}{l|}{Swahili} &
  \multicolumn{3}{c}{1.11 \scriptsize{± 0.23}} &
  \multicolumn{3}{c}{3.96 \scriptsize{± 0.11}} &
  \multicolumn{3}{c}{6.48 \scriptsize{± 0.24}} &
  \multicolumn{3}{c|}{3.85 \scriptsize{± 0.19}} &
  \multicolumn{3}{c}{0.65 \scriptsize{± 0.07}} &
  \multicolumn{3}{c}{2.29 \scriptsize{± 0.06}} &
  \multicolumn{3}{c}{3.67 \scriptsize{± 0.08}} &
  \multicolumn{3}{c|}{2.20 \scriptsize{± 0.02}} &
  \multicolumn{3}{c}{0.61 \scriptsize{± 0.18}} &
  \multicolumn{3}{c}{2.89 \scriptsize{± 0.32}} &
  \multicolumn{3}{c}{5.10 \scriptsize{± 0.18}} &
  \multicolumn{3}{c|}{2.87 \scriptsize{± 0.21}} &
  \multicolumn{3}{c}{0.59 \scriptsize{± 0.03}} &
  \multicolumn{3}{c}{2.21 \scriptsize{± 0.06}} &
  \multicolumn{3}{c}{3.49 \scriptsize{± 0.10}} &
  \multicolumn{3}{c}{2.10 \scriptsize{± 0.06}} \\
\multicolumn{1}{l|}{Japanese} &
  \multicolumn{3}{c}{21.36 \scriptsize{± 0.49}} &
  \multicolumn{3}{c}{51.46 \scriptsize{± 0.24}} &
  \multicolumn{3}{c}{65.64 \scriptsize{± 0.38}} &
  \multicolumn{3}{c|}{46.15 \scriptsize{± 0.22}} &
  \multicolumn{3}{c}{5.59 \scriptsize{± 0.24}} &
  \multicolumn{3}{c}{21.87 \scriptsize{± 0.48}} &
  \multicolumn{3}{c}{33.82 \scriptsize{± 0.20}} &
  \multicolumn{3}{c|}{20.42 \scriptsize{± 0.30}} &
  \multicolumn{3}{c}{18.35 \scriptsize{± 0.52}} &
  \multicolumn{3}{c}{45.76 \scriptsize{± 0.96}} &
  \multicolumn{3}{c}{59.78 \scriptsize{± 0.80}} &
  \multicolumn{3}{c|}{41.30 \scriptsize{± 0.68}} &
  \multicolumn{3}{c}{5.76 \scriptsize{± 0.18}} &
  \multicolumn{3}{c}{22.74 \scriptsize{± 0.30}} &
  \multicolumn{3}{c}{34.45 \scriptsize{± 0.05}} &
  \multicolumn{3}{c}{20.98 \scriptsize{± 0.10}} \\
\multicolumn{1}{l|}{Turkish} &
  \multicolumn{3}{c}{17.04 \scriptsize{± 0.40}} &
  \multicolumn{3}{c}{42.65 \scriptsize{± 0.63}} &
  \multicolumn{3}{c}{56.80 \scriptsize{± 0.45}} &
  \multicolumn{3}{c|}{38.84 \scriptsize{± 0.47}} &
  \multicolumn{3}{c}{4.42 \scriptsize{± 0.23}} &
  \multicolumn{3}{c}{17.22 \scriptsize{± 0.32}} &
  \multicolumn{3}{c}{27.15 \scriptsize{± 0.51}} &
  \multicolumn{3}{c|}{16.27 \scriptsize{± 0.33}} &
  \multicolumn{3}{c}{12.76 \scriptsize{± 0.15}} &
  \multicolumn{3}{c}{35.07 \scriptsize{± 0.38}} &
  \multicolumn{3}{c}{47.96 \scriptsize{± 0.51}} &
  \multicolumn{3}{c|}{31.93 \scriptsize{± 0.34}} &
  \multicolumn{3}{c}{4.52 \scriptsize{± 0.16}} &
  \multicolumn{3}{c}{17.64 \scriptsize{± 0.22}} &
  \multicolumn{3}{c}{27.81 \scriptsize{± 0.29}} &
  \multicolumn{3}{c}{16.66 \scriptsize{± 0.06}} \\ \hline
 &
  \multicolumn{48}{c}{ClothoV2} \\ \hline
\multicolumn{1}{l|}{English} &
  \multicolumn{3}{c}{13.21 \scriptsize{± 3.16}} &
  \multicolumn{3}{c}{33.63 \scriptsize{± 6.91}} &
  \multicolumn{3}{c}{45.67 \scriptsize{± 8.51}} &
  \multicolumn{3}{c|}{30.84 \scriptsize{± 6.19}} &
  \multicolumn{3}{c}{2.59  \scriptsize{± 0.74}} &
  \multicolumn{3}{c}{10.07 \scriptsize{± 2.87}} &
  \multicolumn{3}{c}{16.62 \scriptsize{± 4.68}} &
  \multicolumn{3}{c|}{9.76 \scriptsize{± 2.76}} &
  \multicolumn{3}{c}{15.67 \scriptsize{± 0.50}} &
  \multicolumn{3}{c}{38.46 \scriptsize{± 0.44}} &
  \multicolumn{3}{c}{51.25 \scriptsize{± 0.06}} &
  \multicolumn{3}{c|}{35.13 \scriptsize{± 0.31}} &
  \multicolumn{3}{c}{3.28 \scriptsize{± 0.17}} &
  \multicolumn{3}{c}{13.57 \scriptsize{± 0.14}} &
  \multicolumn{3}{c}{21.51 \scriptsize{± 0.05}} &
  \multicolumn{3}{c}{12.79 \scriptsize{± 0.08}} \\
\multicolumn{1}{l|}{Portuguese} &
  \multicolumn{3}{c}{11.16 \scriptsize{± 0.08}} &
  \multicolumn{3}{c}{30.27 \scriptsize{± 0.56}} &
  \multicolumn{3}{c}{41.59 \scriptsize{± 0.48}} &
  \multicolumn{3}{c|}{27.68 \scriptsize{± 0.33}} &
  \multicolumn{3}{c}{2.26 \scriptsize{± 0.17}} &
  \multicolumn{3}{c}{9.47 \scriptsize{± 0.43}} &
  \multicolumn{3}{c}{15.82 \scriptsize{± 0.50}} &
  \multicolumn{3}{c|}{9.18 \scriptsize{± 0.31}} &
  \multicolumn{3}{c}{11.17 \scriptsize{± 0.20}} &
  \multicolumn{3}{c}{30.45 \scriptsize{± 0.28}} &
  \multicolumn{3}{c}{41.94 \scriptsize{± 0.72}} &
  \multicolumn{3}{c|}{27.85 \scriptsize{± 0.40}} &
  \multicolumn{3}{c}{2.64 \scriptsize{± 0.10}} &
  \multicolumn{3}{c}{10.92 \scriptsize{± 0.06}} &
  \multicolumn{3}{c}{17.91 \scriptsize{± 0.10}} &
  \multicolumn{3}{c}{10.49 \scriptsize{± 0.06}} \\
\multicolumn{1}{l|}{Spanish} &
  \multicolumn{3}{c}{11.67 \scriptsize{± 0.24}} &
  \multicolumn{3}{c}{30.95 \scriptsize{± 0.33}} &
  \multicolumn{3}{c}{43.00 \scriptsize{± 0.63}} &
  \multicolumn{3}{c|}{28.54 \scriptsize{± 0.21}} &
  \multicolumn{3}{c}{2.30 \scriptsize{± 0.11}} &
  \multicolumn{3}{c}{9.72 \scriptsize{± 0.37}} &
  \multicolumn{3}{c}{15.67 \scriptsize{± 0.16}} &
  \multicolumn{3}{c|}{9.23 \scriptsize{± 0.19}} &
  \multicolumn{3}{c}{11.67 \scriptsize{± 0.40}} &
  \multicolumn{3}{c}{31.33 \scriptsize{± 0.25}} &
  \multicolumn{3}{c}{42.58 \scriptsize{± 0.39}} &
  \multicolumn{3}{c|}{28.53 \scriptsize{± 0.23}} &
  \multicolumn{3}{c}{2.63 \scriptsize{± 0.18}} &
  \multicolumn{3}{c}{10.87 \scriptsize{± 0.18}} &
  \multicolumn{3}{c}{17.66 \scriptsize{± 0.46}} &
  \multicolumn{3}{c}{10.39 \scriptsize{± 0.15}} \\
\multicolumn{1}{l|}{French} &
  \multicolumn{3}{c}{10.30 \scriptsize{± 0.14}} &
  \multicolumn{3}{c}{28.72 \scriptsize{± 0.06}} &
  \multicolumn{3}{c}{40.27 \scriptsize{± 0.13}} &
  \multicolumn{3}{c|}{26.43 \scriptsize{± 0.06}} &
  \multicolumn{3}{c}{2.28 \scriptsize{± 0.18}} &
  \multicolumn{3}{c}{9.63 \scriptsize{± 0.39}} &
  \multicolumn{3}{c}{15.67 \scriptsize{± 0.15}} &
  \multicolumn{3}{c|}{9.20 \scriptsize{± 0.16}} &
  \multicolumn{3}{c}{10.45 \scriptsize{± 0.74}} &
  \multicolumn{3}{c}{28.07 \scriptsize{± 1.20}} &
  \multicolumn{3}{c}{39.78 \scriptsize{± 0.82}} &
  \multicolumn{3}{c|}{26.10 \scriptsize{± 0.80}} &
  \multicolumn{3}{c}{2.56 \scriptsize{± 0.21}} &
  \multicolumn{3}{c}{10.68 \scriptsize{± 0.32}} &
  \multicolumn{3}{c}{17.36 \scriptsize{± 0.12}} &
  \multicolumn{3}{c}{10.20 \scriptsize{± 0.19}} \\
\multicolumn{1}{l|}{Russian} &
  \multicolumn{3}{c}{9.53 \scriptsize{± 0.21}} &
  \multicolumn{3}{c}{26.88 \scriptsize{± 0.35}} &
  \multicolumn{3}{c}{38.57 \scriptsize{± 0.23}} &
  \multicolumn{3}{c|}{25.00 \scriptsize{± 0.15}} &
  \multicolumn{3}{c}{2.11 \scriptsize{± 0.10}} &
  \multicolumn{3}{c}{8.29 \scriptsize{± 0.24}} &
  \multicolumn{3}{c}{13.70 \scriptsize{± 0.76}} &
  \multicolumn{3}{c|}{8.03 \scriptsize{± 0.35}} &
  \multicolumn{3}{c}{10.09 \scriptsize{± 0.52}} &
  \multicolumn{3}{c}{27.71 \scriptsize{± 0.76}} &
  \multicolumn{3}{c}{39.13 \scriptsize{± 0.77}} &
  \multicolumn{3}{c|}{25.64 \scriptsize{± 0.66}} &
  \multicolumn{3}{c}{2.54 \scriptsize{± 0.09}} &
  \multicolumn{3}{c}{9.79 \scriptsize{± 0.31}} &
  \multicolumn{3}{c}{15.85 \scriptsize{± 0.29}} &
  \multicolumn{3}{c}{9.40 \scriptsize{± 0.20}} \\
\multicolumn{1}{l|}{Arabic} &
  \multicolumn{3}{c}{7.90 \scriptsize{± 0.19}} &
  \multicolumn{3}{c}{23.31 \scriptsize{± 0.52}} &
  \multicolumn{3}{c}{34.68 \scriptsize{± 0.03}} &
  \multicolumn{3}{c|}{21.96 \scriptsize{± 0.20}} &
  \multicolumn{3}{c}{1.71 \scriptsize{± 0.26}} &
  \multicolumn{3}{c}{7.06 \scriptsize{± 0.29}} &
  \multicolumn{3}{c}{12.02 \scriptsize{± 0.34}} &
  \multicolumn{3}{c|}{6.93 \scriptsize{± 0.29}} &
  \multicolumn{3}{c}{7.99 \scriptsize{± 0.10}} &
  \multicolumn{3}{c}{23.64 \scriptsize{± 0.12}} &
  \multicolumn{3}{c}{34.04 \scriptsize{± 0.28}} &
  \multicolumn{3}{c|}{21.89 \scriptsize{± 0.03}} &
  \multicolumn{3}{c}{2.04 \scriptsize{± 0.02}} &
  \multicolumn{3}{c}{8.22 \scriptsize{± 0.14}} &
  \multicolumn{3}{c}{13.47 \scriptsize{± 0.21}} &
  \multicolumn{3}{c}{7.91 \scriptsize{± 0.12}} \\
\multicolumn{1}{l|}{Hindi} &
  \multicolumn{3}{c}{6.55 \scriptsize{± 0.27}} &
  \multicolumn{3}{c}{19.04 \scriptsize{± 0.26}} &
  \multicolumn{3}{c}{28.74 \scriptsize{± 0.44}} &
  \multicolumn{3}{c|}{18.11 \scriptsize{± 0.29}} &
  \multicolumn{3}{c}{1.56 \scriptsize{± 0.12}} &
  \multicolumn{3}{c}{6.07 \scriptsize{± 0.20}} &
  \multicolumn{3}{c}{10.26 \scriptsize{± 0.32}} &
  \multicolumn{3}{c|}{5.97 \scriptsize{± 0.10}} &
  \multicolumn{3}{c}{6.81 \scriptsize{± 0.66}} &
  \multicolumn{3}{c}{19.51 \scriptsize{± 0.89}} &
  \multicolumn{3}{c}{28.32 \scriptsize{± 0.85}} &
  \multicolumn{3}{c|}{18.22 \scriptsize{± 0.79}} &
  \multicolumn{3}{c}{1.85 \scriptsize{± 0.09}} &
  \multicolumn{3}{c}{6.98 \scriptsize{± 0.32}} &
  \multicolumn{3}{c}{11.46 \scriptsize{± 0.07}} &
  \multicolumn{3}{c}{6.76 \scriptsize{± 0.11}}
  \\
\multicolumn{1}{l|}{German} &
  \multicolumn{3}{c}{10.95 \scriptsize{± 0.25}} &
  \multicolumn{3}{c}{29.53 \scriptsize{± 0.63}} &
  \multicolumn{3}{c}{40.91 \scriptsize{± 0.83}} &
  \multicolumn{3}{c|}{27.13 \scriptsize{± 0.57}} &
 \multicolumn{3}{c}{2.62 \scriptsize{± 0.09}} &
  \multicolumn{3}{c}{10.11 \scriptsize{± 0.22}} &
  \multicolumn{3}{c}{16.54 \scriptsize{± 0.37}} &
  \multicolumn{3}{c|}{9.76 \scriptsize{± 0.21}} &
  \multicolumn{3}{c}{11.22 \scriptsize{± 0.39}} &
  \multicolumn{3}{c}{30.31 \scriptsize{± 0.74}} &
  \multicolumn{3}{c}{41.44 \scriptsize{± 1.40}} &
  \multicolumn{3}{c|}{27.66 \scriptsize{± 0.81}} &
  \multicolumn{3}{c}{2.72 \scriptsize{± 0.07}} &
  \multicolumn{3}{c}{11.12 \scriptsize{± 0.31}} &
  \multicolumn{3}{c}{18.30 \scriptsize{± 0.48}} &
  \multicolumn{3}{c}{10.72 \scriptsize{± 0.27}} \\
\multicolumn{1}{l|}{Chinese(zh)} &
  \multicolumn{3}{c}{9.90 \scriptsize{± 0.35}} &
  \multicolumn{3}{c}{27.41 \scriptsize{± 0.56}} &
  \multicolumn{3}{c}{39.20 \scriptsize{± 0.55}} &
  \multicolumn{3}{c|}{25.51 \scriptsize{± 0.31}} &
  \multicolumn{3}{c}{2.26 \scriptsize{± 0.05}} &
  \multicolumn{3}{c}{8.76 \scriptsize{± 0.24}} &
  \multicolumn{3}{c}{14.25 \scriptsize{± 0.20}} &
  \multicolumn{3}{c|}{8.43 \scriptsize{± 0.13}} &
  \multicolumn{3}{c}{9.98 \scriptsize{± 0.15}} &
  \multicolumn{3}{c}{27.28 \scriptsize{± 0.14}} &
  \multicolumn{3}{c}{38.59 \scriptsize{± 0.80}} &
  \multicolumn{3}{c|}{25.28 \scriptsize{± 0.36}} &
  \multicolumn{3}{c}{2.48 \scriptsize{± 0.10}} &
  \multicolumn{3}{c}{9.71 \scriptsize{± 0.17}} &
  \multicolumn{3}{c}{15.53 \scriptsize{± 0.21}} &
  \multicolumn{3}{c}{9.24 \scriptsize{± 0.14}} \\
\multicolumn{1}{l|}{Swahili} &
  \multicolumn{3}{c}{0.78 \scriptsize{± 0.06}} &
  \multicolumn{3}{c}{3.44 \scriptsize{± 0.26}} &
  \multicolumn{3}{c}{5.61 \scriptsize{± 0.12}} &
  \multicolumn{3}{c|}{3.28 \scriptsize{± 0.11}} &
  \multicolumn{3}{c}{0.42 \scriptsize{± 0.05}} &
  \multicolumn{3}{c}{1.54 \scriptsize{± 0.07}} &
  \multicolumn{3}{c}{2.43 \scriptsize{± 0.05}} &
  \multicolumn{3}{c|}{1.46 \scriptsize{± 0.02}} &
  \multicolumn{3}{c}{0.75 \scriptsize{± 0.10}} &
  \multicolumn{3}{c}{2.98 \scriptsize{± 0.07}} &
  \multicolumn{3}{c}{5.20 \scriptsize{± 0.38}} &
  \multicolumn{3}{c|}{2.98 \scriptsize{± 0.10}} &
  \multicolumn{3}{c}{0.55 \scriptsize{± 0.01}} &
  \multicolumn{3}{c}{1.65 \scriptsize{± 0.12}} &
  \multicolumn{3}{c}{2.54 \scriptsize{± 0.23}} &
  \multicolumn{3}{c}{1.58 \scriptsize{± 0.12}} \\
\multicolumn{1}{l|}{Japanese} &
  \multicolumn{3}{c}{10.49 \scriptsize{± 0.35}} &
  \multicolumn{3}{c}{28.92 \scriptsize{± 0.41}} &
  \multicolumn{3}{c}{40.76 \scriptsize{± 0.76}} &
  \multicolumn{3}{c|}{26.72 \scriptsize{± 0.44}} &
  \multicolumn{3}{c}{2.06 \scriptsize{± 0.14}} &
  \multicolumn{3}{c}{8.86 \scriptsize{± 0.30}} &
  \multicolumn{3}{c}{14.51 \scriptsize{± 0.43}} &
  \multicolumn{3}{c|}{8.48 \scriptsize{± 0.25}} &
  \multicolumn{3}{c}{10.48 \scriptsize{± 0.23}} &
  \multicolumn{3}{c}{28.87 \scriptsize{± 0.26}} &
  \multicolumn{3}{c}{40.19 \scriptsize{± 0.44}} &
  \multicolumn{3}{c|}{26.51 \scriptsize{± 0.31}} &
  \multicolumn{3}{c}{2.35 \scriptsize{± 0.20}} &
  \multicolumn{3}{c}{9.80 \scriptsize{± 0.22}} &
  \multicolumn{3}{c}{16.03 \scriptsize{± 0.16}} &
  \multicolumn{3}{c}{9.39 \scriptsize{± 0.17}} \\
\multicolumn{1}{l|}{Turkish} &
  \multicolumn{3}{c}{8.60 \scriptsize{± 0.48}} &
  \multicolumn{3}{c}{24.53 \scriptsize{± 0.82}} &
  \multicolumn{3}{c}{35.75 \scriptsize{± 0.55}} &
  \multicolumn{3}{c|}{22.96 \scriptsize{± 0.60}} &
  \multicolumn{3}{c}{1.74 \scriptsize{± 0.09}} &
  \multicolumn{3}{c}{7.04 \scriptsize{± 0.09}} &
  \multicolumn{3}{c}{12.13 \scriptsize{± 0.00}} &
  \multicolumn{3}{c|}{6.97 \scriptsize{± 0.06}} &
  \multicolumn{3}{c}{8.45 \scriptsize{± 0.06}} &
  \multicolumn{3}{c}{24.02 \scriptsize{± 0.50}} &
  \multicolumn{3}{c}{34.61 \scriptsize{± 0.75}} &
  \multicolumn{3}{c|}{22.36 \scriptsize{± 0.41}} &
  \multicolumn{3}{c}{2.14 \scriptsize{± 0.07}} &
  \multicolumn{3}{c}{8.06 \scriptsize{± 0.24}} &
  \multicolumn{3}{c}{13.56 \scriptsize{± 0.19}} &
  \multicolumn{3}{c}{7.92 \scriptsize{± 0.13}} \\ \hline
\end{tabular}%
}
\label{tab:retrieval_expanded_resources}
\end{table*}

For the classification task, comparing both datasets through Table~\ref{tab:classification_expanded_resources}, results are generally improved when using more resources with more varied data. This shows that this task benefits from more robust training. Despite benefiting from a more robust structure, the presented model continues to demonstrate excellent results with a modest computational budget but has the capacity to achieve better results with greater computational power and data availability during training. In this structure, we continued to train the model only in English; the other languages improved due to emergent learning.

\begin{table}[!htb]
\caption{Classification task results, evaluated on ESC-50 and UrbanSounds8K datasets for expanded resources experiments cross the 12~evaluated languages.}
\resizebox{0.7\columnwidth}{!}{%
\begin{tabular}{lccrccrccrccr}
\hline
& \multicolumn{3}{c}{OB Model} & \multicolumn{3}{l}{ER Model} & \multicolumn{3}{c}{OB Model} & \multicolumn{3}{c}{ER Model} \\ \hline
            & \multicolumn{6}{c|}{ESC-50}                              & \multicolumn{6}{c}{UrbanSounds8K}   \\ \hline
Language    & \multicolumn{12}{c}{Accuracy (\%)}                                                                  \\ \hline
English     & \multicolumn{3}{c}{91.35 \scriptsize{± 0.69}} & \multicolumn{3}{c|}{93.50 \scriptsize{± 0.18}} &  \multicolumn{3}{c}{74.99 \scriptsize{± 0.55}} & \multicolumn{3}{c}{78.33 \scriptsize{± 2.75}} \\
Portuguese  & \multicolumn{3}{c}{80.92 \scriptsize{± 0.98}} & \multicolumn{3}{c|}{81.70 \scriptsize{± 0.88}} &  \multicolumn{3}{c}{71.38 \scriptsize{± 1.35}} & \multicolumn{3}{c}{68.15 \scriptsize{± 1.63}} \\
Spanish     & \multicolumn{3}{c}{85.60 \scriptsize{± 0.78}} & \multicolumn{3}{c|}{87.65 \scriptsize{± 0.43}} &  \multicolumn{3}{c}{71.02 \scriptsize{± 0.67}} & \multicolumn{3}{c}{72.72 \scriptsize{± 1.50}} \\
French      & \multicolumn{3}{c}{82.45 \scriptsize{± 0.49}} & \multicolumn{3}{c|}{83.85 \scriptsize{± 0.17}} &  \multicolumn{3}{c}{67.72 \scriptsize{± 1.12}} & \multicolumn{3}{c}{70.47 \scriptsize{± 1.82}} \\
Russian     & \multicolumn{3}{c}{78.30 \scriptsize{± 1.08}} & \multicolumn{3}{c|}{80.62 \scriptsize{± 1.36}} &  \multicolumn{3}{c}{67.16 \scriptsize{± 1.32}} & \multicolumn{3}{c}{69.32 \scriptsize{± 0.22}} \\
Arabic      & \multicolumn{3}{c}{63.65 \scriptsize{± 1.09}} & \multicolumn{3}{c|}{63.67 \scriptsize{± 0.77}} &  \multicolumn{3}{c}{63.77 \scriptsize{± 0.60}} & \multicolumn{3}{c}{62.40 \scriptsize{± 1.48}} \\
Hindi       & \multicolumn{3}{c}{60.12 \scriptsize{± 0.83}} & \multicolumn{3}{c|}{62.57 \scriptsize{± 0.88}} &  \multicolumn{3}{c}{58.11 \scriptsize{± 1.15}} & \multicolumn{3}{c}{56.55 \scriptsize{± 0.78}} \\
German      & \multicolumn{3}{c}{75.92 \scriptsize{± 1.08}} & \multicolumn{3}{c|}{78.37 \scriptsize{± 0.64}} &  \multicolumn{3}{c}{72.38 \scriptsize{± 0.10}} & \multicolumn{3}{c}{74.44 \scriptsize{± 1.24}} \\
Chinese(zh) & \multicolumn{3}{c}{81.23 \scriptsize{± 1.07}} & \multicolumn{3}{c|}{82.80 \scriptsize{± 0.93}} &  \multicolumn{3}{c}{67.22 \scriptsize{± 1.15}} & \multicolumn{3}{c}{68.50 \scriptsize{± 0.59}} \\
Swahili     & \multicolumn{3}{c}{20.48 \scriptsize{± 2.17}} & \multicolumn{3}{c|}{21.80 \scriptsize{± 0.22}} &  \multicolumn{3}{c}{38.35 \scriptsize{± 1.97}} & \multicolumn{3}{c}{36.28 \scriptsize{± 3.61}} \\
Japanese    & \multicolumn{3}{c}{81.78 \scriptsize{± 0.75}} & \multicolumn{3}{c|}{85.83 \scriptsize{± 1.46}} &  \multicolumn{3}{c}{73.03 \scriptsize{± 0.74}} & \multicolumn{3}{c}{72.18 \scriptsize{± 1.88}} \\
Turkish     & \multicolumn{3}{c}{69.85 \scriptsize{± 1.97}} & \multicolumn{3}{c|}{72.92 \scriptsize{± 0.81}} &  \multicolumn{3}{c}{63.91 \scriptsize{± 0.82}} & \multicolumn{3}{c}{64.52 \scriptsize{± 0.82}} \\ \hline
\end{tabular}%
}
\label{tab:classification_expanded_resources}
\end{table}

\subsection{Qualitative Analysis}
\label{sec:appendix_qualitative_analysis}

\begin{figure}[!htb]
    \includegraphics[width=0.95\linewidth]{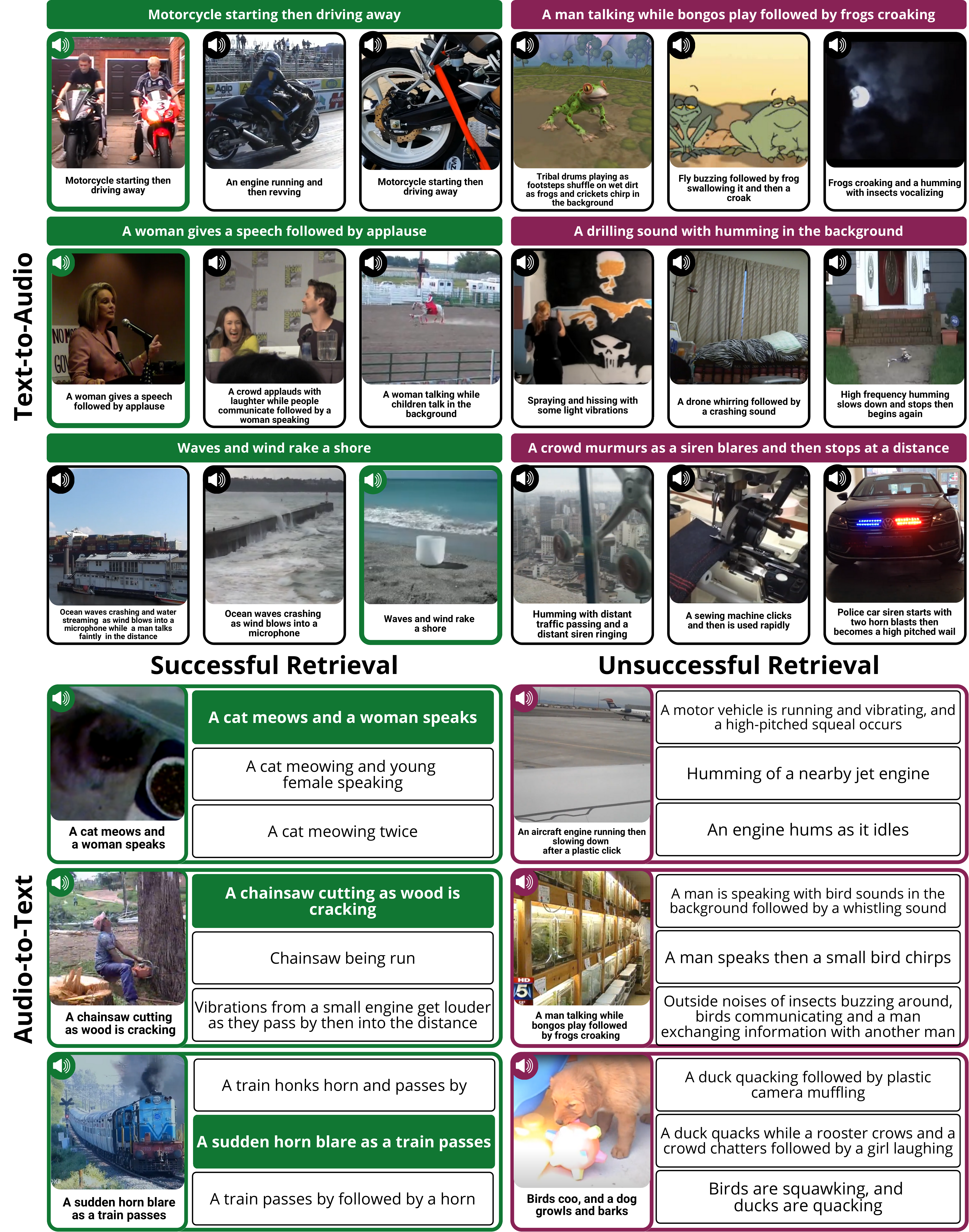}
    \caption{Examples of successful (left) and unsuccessful (right) text-to-audio retrieval, shown at the top of the image. The green-highlighted image and text indicate the ground-truth audio clip retrieved within the top 3 results for the given text query. For unsuccessful text queries, we present all the returned audio clips in image format. At the bottom, successful (left) and unsuccessful (right) audio-to-text retrieval are shown. The green-highlighted text indicates the ground-truth text description retrieved for the given audio query. For unsuccessful audio queries, all the returned text results are presented.}
    \label{fig:qualitative_audio_text}
\end{figure}

Figure~\ref{fig:qualitative_audio_text} illustrates successful results of the text-to-audio retrieval task. In each example shown, the ground-truth audio clip corresponding to the input text query is among the top-3 ranked retrieval results. These ground truth matches are highlighted in red. For each retrieved audio clip, we display a representative frame extracted from the corresponding YouTube video segment and the original caption associated with the audio clip. This visualization enables a direct comparison between the textual query, the retrieved audio content (represented by the video frame and caption), and the ground truth.

\begin{figure} [!htb]
    \includegraphics[width=0.95\linewidth]{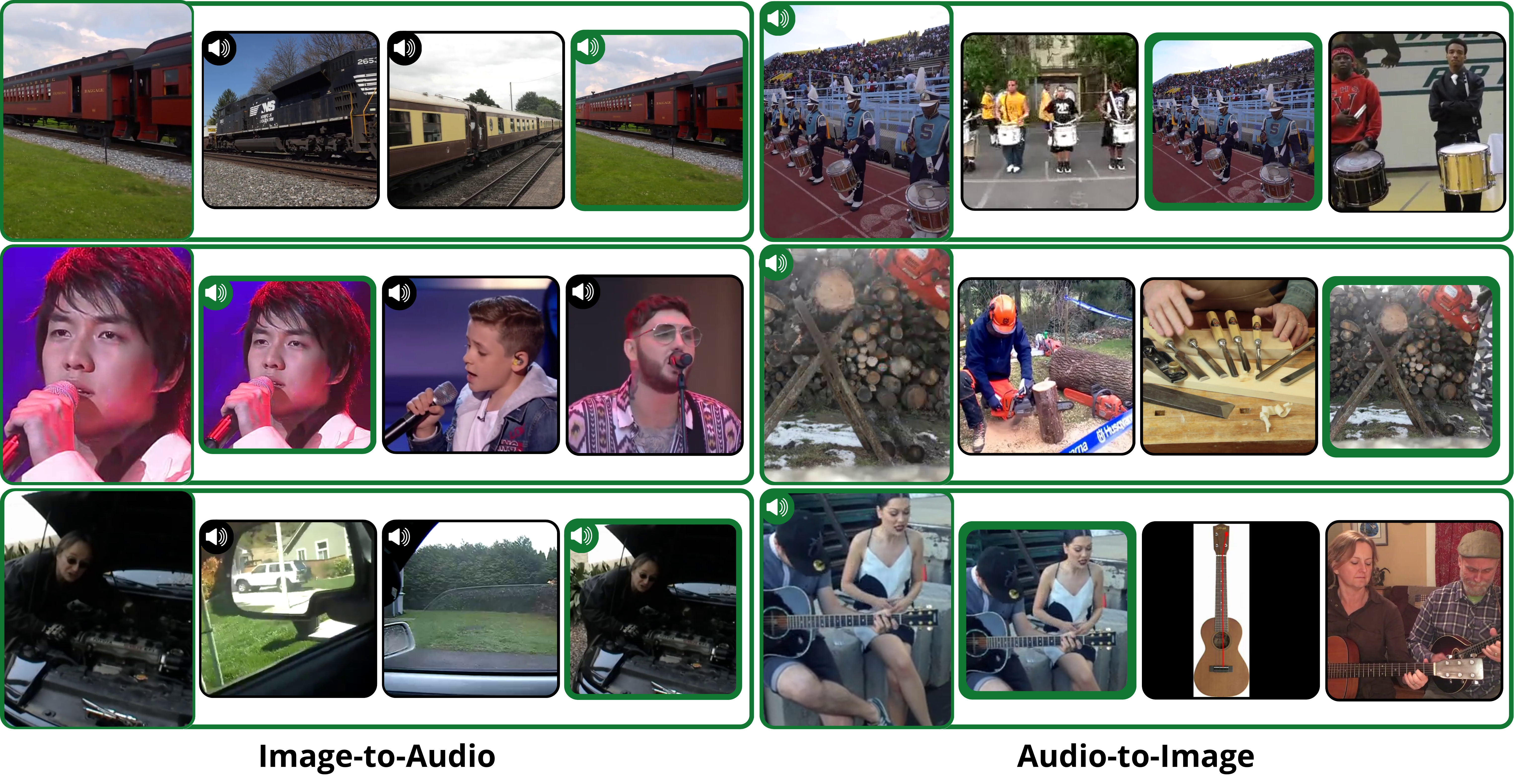}
    \caption{Examples of successful audio-to-image retrieval (left) and image-to-audio retrieval (right). The green-highlighted image shows the ground-truth image retrieved for the given audio~query.}
    \label{fig:qualitative_audio_image}
\end{figure}

To qualitatively evaluate our model's cross-modal retrieval capabilities, we conducted experiments on the AudioCaps test dataset. We present representative examples of successful and unsuccessful retrieval outcomes for both text-to-audio and audio-to-text tasks in Figure~\ref{fig:qualitative_audio_text}. We also investigate the emergent capabilities of our model in audio-to-image and image-to-audio retrieval tasks, demonstrating its ability to implicitly align the audio and image modalities that were not explicitly aligned during training. Specifically, we leveraged the test subset of the \text{VGG-Sound} dataset, extracting audio and the central video frame following the methodology outlined by~\citet{guzhov2022audioclip}. These results are visualized in Figure~\ref{fig:qualitative_audio_image}.

Figure~\ref{fig:qualitative_audio_text}, at the top of the image, on the left side, shows successful audio-to-text retrieval. Given an audio clip as a query, the model retrieves the top-ranked textual descriptions. The retrieved text is displayed, with the ground truth description highlighted in red. To provide context for the audio query, we include a representative frame extracted from the corresponding YouTube video segment. This frame, along with the original text caption associated with the audio, helps to clarify the content of the audio query. At the top of the image, on the right, are examples of unsuccessful text-to-audio retrieval. In these instances, the ground truth audio clip corresponding to the input text query is absent from the top-3 ranked retrieval results, indicating a mismatch between the text and the retrieved audio.

The image illustrates successful audio-to-text retrieval at the bottom left. Given an audio clip as a query, the model retrieves the top-ranked images. The retrieved image is displayed, with the ground truth description highlighted in red. In the bottom right are examples of unsuccessful audio-to-text retrieval. Here, the ground truth textual description corresponding to the audio query is not found within the top-3 ranked retrieval results, indicating a failure to accurately capture the audio's content in the retrieved text. Figure~\ref{fig:qualitative_audio_image} presents successful image-to-audio and audio-to-image retrieval. Given an image or an audio query, the model retrieves the top-ranked images and audio. The retrieved audio is represented by a representative frame from the corresponding video and the associated caption. The ground truth audio is highlighted in green.

\section{Conclusions}
\label{sec:conclusions}

Despite numerous efforts in the literature to develop multimodal and multilingual models, many approaches fail to leverage prior knowledge or to optimize training efficiency effectively. In this work, we proposed CACARA, an architecture and model based on emergent alignment learning that can integrate a new modality into an existing bimodal/multimodal architecture without requiring full retraining. Additionally, our approach enables the multilingual expansion of the newly added modality to all supported languages.

By leveraging emergent alignment, our method simplifies training. It significantly reduces computational costs by eliminating the need to retrain all components while enhancing conceptual complementarity across modalities. Our results demonstrate strong alignment across the integrated modalities, achieving superior R@1 performance compared to most state-of-the-art multimodal models in the literature.
\section*{Limitations}
\label{sec:limitations}

While our model demonstrates the capacity for seamless integration of additional modalities without degradation of performance in existing aligned modalities, the present study's scope was limited to the incorporation of the audio modality. Extending the framework to incorporate additional modalities (e.g., video, thermal, wearable sensor, and depth data) would provide a more comprehensive assessment of the model's scalability and ability to generalize across diverse representational spaces.

Furthermore, our experimental design was constrained to base-level encoder models. A more rigorous evaluation of the model's effectiveness would necessitate an investigation across a spectrum of encoder sizes. This would entail not only increased computational resources (data quantity, batch size, and training duration) but also a systematic exploration of the relationship between model parameterization and performance gains. A larger parameter count would allow one to assess the scalability.

Finally, while our evaluation encompassed 12 languages, demonstrating a degree of multilingual capability, the generalizability of these findings could be further strengthened by expanding the linguistic scope. A more comprehensive evaluation should include languages exhibiting greater typological diversity and, crucially, languages with varying levels of available digital resources. This would allow for a more nuanced understanding of the model's performance in low-resource language settings, which are often underrepresented in current research.

\section*{Ethics Statement}
\label{sec:ethics_statement}

This work focuses on enhancing multimodal and multilingual models by leveraging emergent alignment through implicit learning to reduce computational overhead and enhance accessibility. We fully comply with the terms of use and licensing agreements associated with all datasets used for training, evaluation, or testing our models. This work does not involve human subjects; however, we recognize the ethical and societal responsibilities of deploying such models, including the potential for misuse (e.g., generating harmful or misleading text, audio, or images). Despite efforts to improve a multilingual model's multilingual capabilities, our models may still exhibit biases or underrepresentation of specific languages, cultures, topics, or applications, particularly those with limited data resources, which can be inherited from an already pre-trained language model. While designed for beneficial applications and scientific advancement, these models could be repurposed for unintended uses.

\appendix
\hypertarget{appendixA}{}
\appendixsection{Hyperparameters and Computing Resources}
\label{sec:appendix_hyperparameters}

The hyperparameters used for fine-tuning the general BEAT models are shown in Table~\ref{tab:hyperparameters}. The basic model has a batch size of $64$ and $3$ epochs.

\begin{table*}[!htb]
\caption{Hyperparameters for the Expanded Resources and Basic Model configurations.}
\small
\begin{tabular}{lcc}
\hline
\textbf{Hyperparameter} & \textbf{Expanded Resources} & \textbf{Basic Model} \\ \hline
Batch size               & 110                              & 64                       \\
Maximum text token length& 77 & 77                       \\
Maximum audio length     & 10 seconds &      10 seconds           \\
Optimizer                & Adam & Adam                      \\
Weight decay             & 1e-6  & 1e-6                      \\
Adam $\epsilon$          & 1e-8 & 1e-8                     \\
Adam $\beta$             & [0.9, 0.98]   & [0.9, 0.98]              \\
Learning rate schedule   & CosineWarmupLR & CosineWarmupLR           \\
Maximum learning rate    & 5e-5  & 5e-5                    \\
Minimum learning rate    & 1e-5 &    1e-5                   \\
\# Epochs                & 10                               & 2                        \\ \hline
\end{tabular}
\label{tab:hyperparameters}
\end{table*}

To train the base models, we used a 48GB Quadro RTX 8000 GPU. On average, training took 90 hours to complete. The models with expanded computational resources were trained on an NVIDIA A100 GPU with 80GB, with an average training time of $255$ hours.

\hypertarget{appendixB}{}
\appendixsection{Data Augmentation}
\label{sec:appendix_augumentation}

We applied two data augmentation strategies: Random Truncation and SpecAugment. We conducted a preliminary evaluation of both augmentations. Tables~\ref{tab:augumentation-retrieval} and~\ref{tab:augumentation-classification} show the different combinations of these augmentations for the retrieval and classification tasks.

\begin{itemize}
\item \textbf{Random Truncation (RT)} truncates or pads the audio input to a fixed duration (in this work, we use $10$ seconds). For audio clips shorter than the target length, we applied padding in two stages: random padding with silence at the beginning, followed by additional padding to reach the target duration. For longer clips, we extracted a random segment of the required length from the audio. This method introduces variability by exposing the model to different temporal sections of the same audio during training, reducing overfitting while ensuring consistent input dimensions across the dataset.

\item \textbf{SpecAugment (SpecAug)}~\cite{park2019specaugment} operates directly on the log mel spectrogram of input audio rather than the raw waveform. Initially developed for speech recognition tasks, this method has been successfully adopted for sound event detection and audio classification, as demonstrated in several recent studies~\cite{kong2020panns, gong21b_interspeech, chen2023beats}. The method comprises three main operations: (1) time warping, which deforms the time-series along the time direction, (2) frequency masking, where $f$ consecutive mel frequency channels $[f_0, f_0 + f)$ are masked, with $f$ chosen from a uniform distribution from 0 to the frequency mask parameter $F$, and $f_0$ selected from $[0, \nu - f)$ where $\nu$ is the number of frequency channels, and (3) time masking, where $t$ consecutive time steps $[t_0, t_0 + t)$ are masked, with $t$ chosen uniformly from 0 to the time mask parameter $T$, and $t_0$ selected from $[0, \tau - t)$. An upper bound prevents time masks from exceeding $p$ times the number of time steps. Since the spectrograms are normalized to zero mean, setting masked values to zero is equivalent to setting them to the mean value. For the experiments that used SpecAugment we used $F=48$ for the frequency masking parameter and $T=96$ for the time masking parameter. These parameters control the maximum width of the frequency and time masks, respectively.
\end{itemize}

\hypertarget{appendixC}{}
\appendixsection{Datasets}
\label{sec:appendix_datasets}

This section details the datasets used for training, validation, and testing our model. Our training strategy leverages a combination of large-scale, automatically annotated datasets and smaller, high-quality, human-annotated datasets. This approach allows us to benefit from the breadth of data provided by the larger datasets while also incorporating the precision and accuracy afforded by human labeling. Specifically, we utilize AudioSetCaps, WavCaps, Auto-ACD, AudioCaps, and ClothoV2 as our training data.

The rationale for this selection is to balance data quantity and quality. Datasets such as WavCaps, Auto-ACD, and AudioSetCaps provide substantial amounts of data, which are crucial for training robust and generalizable models. While these datasets are automatically annotated and thus potentially contain some noise, their sheer size compensates for this limitation. Prior work has demonstrated the effectiveness of training on these datasets individually, achieving promising results. Our approach builds upon this by combining them, hypothesizing that the combined data will lead to even better performance.

Complementing these large-scale datasets, we incorporate AudioCaps and ClothoV2. These datasets are meticulously annotated by human labelers, providing a ``gold standard'' of data quality. While smaller than automatically generated datasets, these datasets' high accuracy is essential for refining the model's understanding of complex audio-sound relationships and ensuring accurate caption generation. By training on a combination of high-quality, large-scale datasets, we aim to create a model that is both comprehensive in its understanding of audio and accurate in its descriptions.

\begin{itemize}
    \item \textbf{ESC-50}~\cite{piczak2015esc} (Environmental Sound Classification) comprises $2000$ audio clips, each with a duration of $5$ seconds, distributed across $50$ distinct classes. These classes are grouped into five broader categories: animal sounds, natural soundscapes and water sounds, human non-speech sounds, domestic sounds, and urban noises.  Each of these five categories contains $10$ specific sound classes (a total of $50$), each represented by $40$ audio clips. To facilitate consistent evaluation, the dataset provides predefined splits for 5-fold cross-validation.

    \item \textbf{UrbanSound8K}~\cite{salamon2014dataset} focuses specifically on urban environmental sounds, containing $8732$ labeled sound excerpts under $4$ seconds from $10$ distinct urban sound sources. The sounds include air conditioners, car horns, playing children, dog barking, drilling, engine idling, gunshots, jackhammering, sirens, and street music. The dataset is organized into $10$ folds for cross-validation, making it a standard benchmark for urban sound classification.

    \item \textbf{VGG-Sound}~\cite{Chen2020Vggsound} is an audio-visual dataset containing over $200$K video clips of $10$ seconds each, spanning $309$ distinct sound classes. These classes include musical instruments, human sounds, animal vocalizations, environmental noises, and mechanical sounds, with each class containing $200$ to $1000$ clips. The clips were collected from diverse, unconstrained environments to reflect real-world acoustic conditions. The dataset was curated using a multi-stage verification process that included visual classification, audio validation, and noise filtering, ensuring high-quality, consistent data.
    
    \item \textbf{AudioSet}~\cite{gemmeke2017audio} comprises over $2$M human-labeled $10$-second YouTube video excerpts. It is organized in a hierarchical ontology of $527$ sound classes. While extremely comprehensive, it has an unbalanced distribution, with some classes having significantly more samples than others. The dataset provides both balanced and unbalanced training sets, along with a consistent evaluation set.

    \item \textbf{AudioCaps}~\cite{kim2019audiocaps} builds on AudioSet, containing $46$K audio-caption pairs with varying caption density across splits. The training set includes $38118$ clips with single captions, while the validation and test sets comprise $500$ and $979$ clips, respectively, with five captions each. The dataset's curation process deliberately excluded music categories, visually dependent sounds, and expert knowledge categories. During caption collection, annotators received AudioSet labels as word hints, with video hints available as a last resort. The dataset emphasizes the description of auditory content over visual elements.

    \item \textbf{Clotho}~\cite{drossos2020clotho} represents a focused effort on audio captioning with $4981$ audio samples of $15$ to $30$ seconds in duration, and $24905$ captions total. Drawing from the Freesound platform, the audio samples cover diverse environmental and acoustic content. During data collection, annotators wrote captions based solely on audio signals, without access to visual cues or word tags. The dataset underwent post-processing to remove named entities, speech transcriptions, and words that appeared only once, while retaining natural-language descriptions of sound events, acoustic scenes, and spatial-temporal relationships.

    \item \textbf{MACS}~\cite{irene_martin_morato_2021_5114771} (Multi-annotator Captioned Soundscapes) contains approximately $4000$ audio samples with multiple human annotations per clip. Each audio clip is limited to $10$ seconds. What distinguishes MACS is its use of professional annotators and a structured annotation process that ensures high-quality, consistent captions focused purely on auditory content.

    \item \textbf{WavCaps}~\cite{mei2024wavcaps} represents the largest scale effort with approximately $400,000$ audio-caption pairs sourced from FreeSound, BBC Sound Effects, SoundBible, and AudioSet. What sets it apart is its innovative three-stage processing pipeline. First, it filters out clips shorter than one second and removes repetitive descriptions. Then, it employs ChatGPT to transform raw descriptions into proper captions. Finally, it removes named entities and extremely brief captions. While it is considered weakly labeled due to its automated processing, WavCaps maintains caption quality through this structured approach, making it valuable for large-scale audio-language training.

    \item \textbf{Auto-ACD}~\cite{sun2024auto} is a large-scale audio-language dataset containing $1.5$M audio-caption pairs. Each audio clip is paired with a detailed caption averaging $18$ words, drawn from a vocabulary of approximately $23$K words. The captions encompass comprehensive descriptions of acoustic events, environmental context, and scene settings. The dataset uses audio clips from YouTube videos. It provides rich descriptive text that goes beyond simple sound labels, including detailed acoustic and environmental information.

    \item \textbf{AudioSetCaps}~\cite{bai2024audiosetcaps} comprises $1.9$M audio-caption pairs built upon AudioSet recordings. The dataset provides extensive coverage of audio content through detailed captions that describe not only the primary sound events but also their characteristics and environmental context. The captions are enriched with fine-grained audio information, including spoken language details, speech emotions, musical instruments, and music genres. The dataset maintains high caption quality through a refinement process that ensures accuracy and relevance to the audio content.
\end{itemize}

\begin{acknowledgments}
This project was supported by the Ministry of Science, Technology, and Innovation of Brazil, with resources granted by the Federal Law 8.248 of October 23, 1991, under the PPI-Softex. The project was coordinated by Softex and published as Intelligent agents for mobile platforms based on Cognitive Architecture technology \text{[01245.003479/2024-10]}.

D.A.B.M. is partially funded by FAPESP 2023/05939-5. A.I.F., N.S. are partially funded by Centro de Excel\^encia em Intelig\^encia Artificial (CEIA), da Universidade Federal de Goi\'as (UFG). J.S. is funded by FAPESP 2024/23118-1. G.O.S is funded by FAPESP 2024/07969-1. H.P. is partially funded by CNPq (304836/2022-2). S.A. is partially funded by CNPq (316489/2023-9), and FAPESP (2023/12086-9, 2023/12865-8, 2020/09838-0, 2013/08293-7).
\end{acknowledgments}


\newpage
\bibliographystyle{compling}
\bibliography{COLI_template}

\end{document}